%% file: main.tex
\documentclass[final]{cvpr}

\usepackage{times}
\usepackage{epsfig}
\usepackage{graphicx}
\usepackage{amsmath}
\usepackage{amssymb}

\usepackage{amsfonts}
\usepackage{amsthm}
\usepackage{url}
\usepackage{color}
\usepackage{bbm}
\usepackage{tabularx}
\usepackage{xspace}
\usepackage{caption}
\usepackage{booktabs}
\usepackage{microtype}
\usepackage{wrapfig}
\usepackage{adjustbox}
\usepackage[outline]{contour}
\makeatletter
\@namedef{ver@everyshi.sty}{}
\makeatother
\usepackage{tikz}
\usepackage{bbm}
\usepackage{mathtools}
\usepackage{subcaption}
\usepackage{float}

\usepackage{caption}
\newcommand{\D}{\mathrm{d}}

\usepackage[pagebackref=true,breaklinks=true,colorlinks,bookmarks=false]{hyperref}

\def\cvprPaperID{2060} %
\def\confYear{CVPR 2021}

\begin{document}

\title{PhySG: Inverse Rendering with Spherical Gaussians for\\ Physics-based Material Editing and Relighting}

\author{Kai Zhang$^*$
\and 
Fujun Luan$^*$
\and
Qianqian Wang
\and
Kavita Bala
\and
Noah Snavely
\and \\
Cornell University
}

\maketitle

\begin{abstract}
We present PhySG, an end-to-end inverse rendering pipeline that includes a fully differentiable renderer and can reconstruct geometry, materials, and illumination from scratch from a set of RGB input images. 
Our framework represents specular BRDFs and environmental illumination using mixtures of spherical Gaussians, and represents geometry as a signed distance function parameterized as a Multi-Layer Perceptron. 
The use of spherical Gaussians allows us to efficiently solve for approximate light transport, and our method works 
on scenes with challenging non-Lambertian reflectance captured under natural, static illumination. We demonstrate, with both synthetic and real data, that our reconstructions not only enable rendering of novel viewpoints, but also physics-based appearance editing of materials and illumination.
\end{abstract}

\renewcommand{\thefootnote}{\fnsymbol{footnote}}
\footnotetext[1]{Authors contributed equally to this work.}
\footnotetext[2]{Project page: \url{https://kai-46.github.io/PhySG-website/}.}

\input{defs}

\input{sections/math_macros}

\input{sections/01-intro}
\input{sections/02-related}

\input{sections/03-method}
\input{sections/04-experiment}

\input{sections/05-conclusion}

{\small
\bibliographystyle{ieee_fullname}
\bibliography{refs}
}

\end{document}

%% file: defs.tex
\newcommand{\kz}[1]{\noindent {\color{orange} {\bf Kai TODO:} {#1}}}
\newcommand{\gr}[1]{\noindent {\color{blue} {\bf Qianqian TODO:} {#1}}}
\newcommand{\fl}[1]{\noindent {\color{cyan} {\bf Fujun TODO:} {#1}}}
\newcommand{\ns}[1]{\noindent {\color{purple} [{\bf Noah TODO:} {#1}]}}
\newcommand{\kb}[1]{{\color{red}[{\bf KB: }#1]}}

%% file: sections/math_macros.tex
\newcommand{\viewdir}{{\boldsymbol{\omega}_o}}
\newcommand{\lightdir}{{\boldsymbol{\omega}_i}}
\newcommand{\location}{{\mathbf{x}}}
\newcommand{\normaldir}{{\mathbf{n}}}

\newcommand{\lightsgsharp}{\lambda}
\newcommand{\lightsgdir}{\boldsymbol{\xi}}
\newcommand{\lightsgamp}{\boldsymbol{\mu}}

\newcommand{\brdfsgsharp}{\lambda}
\newcommand{\brdfsgdir}{\boldsymbol{\xi}}
\newcommand{\brdfsgamp}{\boldsymbol{\mu}}

\newenvironment{packed_itemize}{
\begin{list}{\labelitemi}{\leftmargin=2em}
\vspace{-6pt}
 \setlength{\itemsep}{0pt}
 \setlength{\parskip}{0pt}
 \setlength{\parsep}{0pt}
}{\end{list}}

%% file: sections/01-intro.tex
\section{Introduction}
Vision as inverse graphics has long been an intriguing concept. Solving inverse rendering problems, i.e., recovering shape, material and lighting from images, has thus been a long-standing goal. Recently, 
neural rendering methods~\cite{tewari2020state,yariv2020multiview,mildenhall2020nerf,niemeyer2020differentiable, martin2020nerf,kaizhang2020, li2020crowdsampling, oechsle2020learning,  meshry2019neural, sitzmann2019scene, sitzmann2019deepvoxels, park2020seeing, azinovic2019inverse,thies2019deferred}, have drawn significant attention 
due to their remarkable success in 
a range of problems, including shape reconstruction, novel view synthesis, non-physically-based relighting, and surface reflectance map estimation. These neural rendering methods adopt scene representations that are either physical, neural, or a mixture of both, along with a neural-network-based renderer. 
Methods that reconstruct textures or radiance fields~\cite{martin2020nerf, yariv2020multiview, niemeyer2020differentiable} work well for the task of interpolating novel views, but do not factorize appearance into 
lighting and materials, precluding physically-based appearance manipulation like material editing or relighting. 
\begin{figure}[H]
    \centering
    \includegraphics[width=1.0\columnwidth]{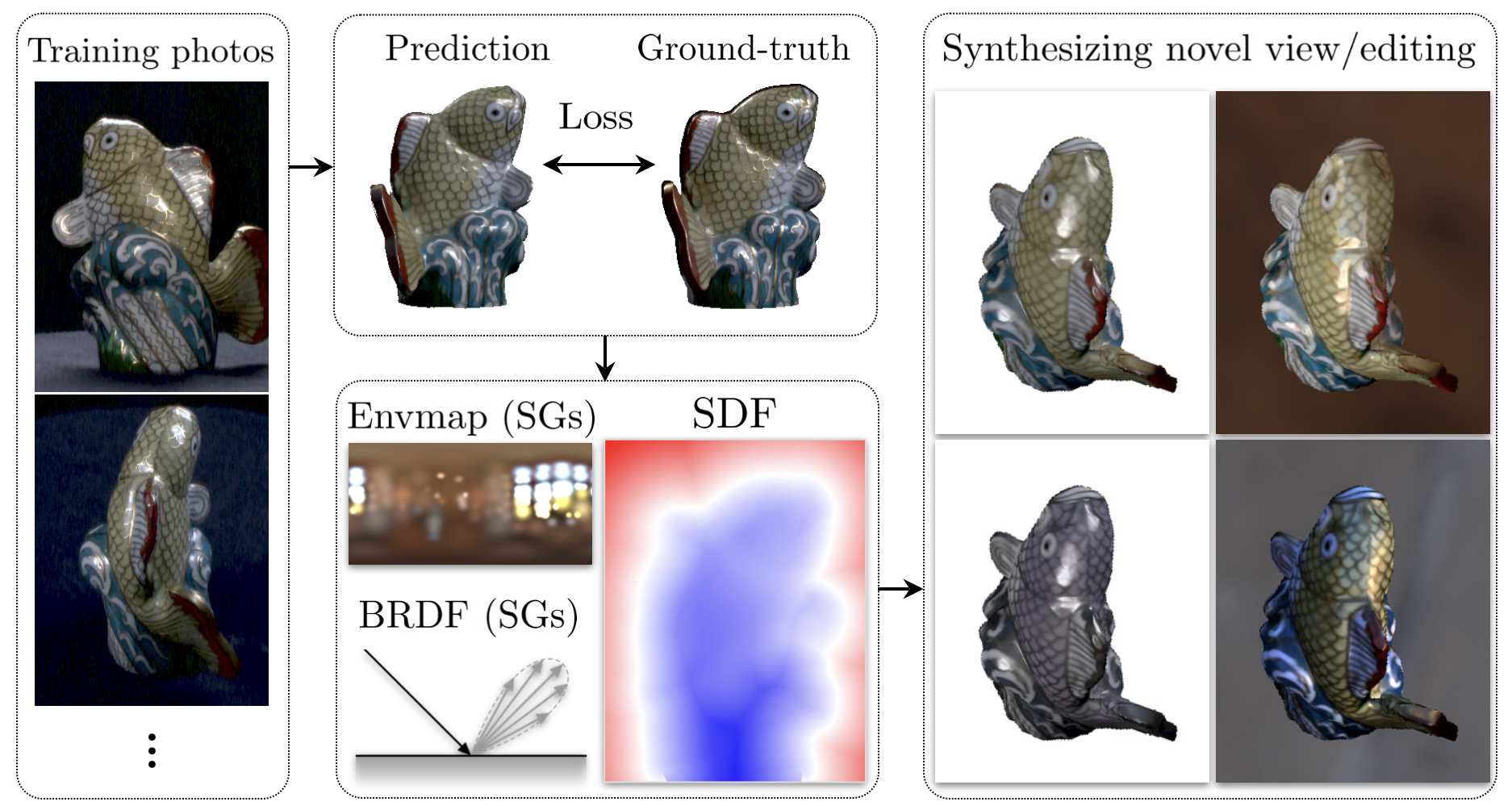}
    \caption{PhySG performs physics-based inverse rendering by taking as input multi-view images of a static glossy object under static natural illumination and jointly optimizes for geometry (represented by an SDF), material BRDF and environment maps (both represented by a mixture of spherical Gaussians), which can then be used for novel view synthesis, relighting and material editing.
    }
    \label{fig:teaser}
\end{figure}

Prior multi-view inverse rendering methods assume RGBD input~\cite{park2020seeing,azinovic2019inverse} or varying illumination across input images achieved either by co-locating an active flashlight with moving cameras~\cite{bi2020deepreflectance,bi2020deep,Schmitt_2020_CVPR, nam2018practical} or capturing objects on a 
turntable with a fixed camera~\cite{xia2016recovering, dong2014appearance}. Learning-based single-view methods that recover shape, illumination, and material properties  have also been proposed~\cite{li2020inverse,barron2014shape}. 

In this work, we tackle the multi-view inverse rendering problem under the challenging setting of normal RGB input images sharing the same static illumination, without assuming scanned geometry. To this end, we propose \textbf{PhySG}, an end-to-end physically-based differentiable rendering pipeline to jointly estimate lighting, material, geometry and surface normals from posed multi-view images of specular objects. In our pipeline, we represent shape using signed distance functions (SDFs), building on their successful use in recent work~\cite{yariv2020multiview,jiang2020sdfdiff,Park_2019_CVPR,maier2017intrinsic3d,zollhofer2015shading}.
Additionally, a key component of our framework is our use of spherical Gaussians to approximate lighting and specular BRDFs
allowing for efficient approximate  evaluation of light transport~\cite{yan2012accurate}. From 2D images alone, our method jointly reconstructs shape, illumination, and materials
and allows for subsequent physics-based appearance manipulations such as  material editing and relighting.

In summary, our contributions are as follows:
\begin{packed_itemize}
\item PhySG, an end-to-end inverse rendering approach to this problem of jointly estimating lighting, material properties, and geometry 
from multi-view images of glossy objects under static illumination. Our pipeline utilizes spherical Gaussians to approximately and efficiently evaluate the rendering equation in closed form. 

\item Compared to prior neural rendering approaches, we show that PhySG not only generalizes to novel viewpoints, but also enables physically-intuitive material editing and relighting. 
\end{packed_itemize}

\vspace{-2mm}

%% file: sections/02-related.tex
\section{Background}
Our approach lies at the intersection of multiple fields. We briefly review the related prior works below.

\medskip
\noindent\textbf{Neural rendering.} The success of neural rendering~\cite{tewari2020state,mildenhall2020nerf,yariv2020multiview,niemeyer2020differentiable,sitzmann2019scene, sitzmann2019deepvoxels,thies2019deferred} has generated significant excitement. In particular, NeRF~\cite{mildenhall2020nerf} enables photo-realistic novel view synthesis by representing scenes as radiance fields via multi-layer-perceptrons (MLPs) and fitting these to a collection of input views. 
While NeRF represents scenes as volumetric opacity fields, other recent methods like DVR~\cite{niemeyer2020differentiable} and IDR~\cite{yariv2020multiview} are surface-based.
In these three works, appearance is represented by a single MLP that takes a 3D point (and a view direction), and outputs a color.
Hence, their appearance model is essentially a surface light field~\cite{wood2000surface} that treats objects as light sources. 
Such an approach works well for novel view synthesis, but does not disentangle material and lighting, and hence is not suitable for physics-based relighting and material editing. Other approaches learn an appearance space~\cite{meshry2019neural,li2020crowdsampling,martin2020nerf} from Internet photos of landmarks captured under diverse lighting, but are not physics-based and cannot generalize to arbitrary new lighting.

In contrast to such prior work that represents appearance as a single neural network, we model appearance via the physical rendering equation. Our approach can solve challenging inverse rendering problems involving specular or glossy objects under static lighting, and enable physically meaningful editing of lighting and materials.

\medskip
\noindent\textbf{Material and environment estimation.} To estimate material properties, most prior works require scenes to be captured under varying illumination~\cite{bi2020neural,bi2020deepreflectance,bi2020deep,Schmitt_2020_CVPR,nam2018practical,xia2016recovering,dong2014appearance,haber2009relighting}.
They either place the object of interest on a mechanical turntable and capture it with a fixed camera~\cite{xia2016recovering,dong2014appearance}, or move a camera with co-located flashlight to capture a static object from multiple viewpoints~\cite{bi2020neural,bi2020deepreflectance,bi2020deep,Schmitt_2020_CVPR,nam2018practical}. The varying illumination 
yields rich cues for inferring material properties and geometry~\cite{ackermann2015survey}. 
For environment estimation from multi-view images, prior works~\cite{park2020seeing,lombardi2016radiometric} factorize scene appearance into diffuse image and surface reflectance map given high-quality geometry from RGBD sensors. The surface reflectance map entangles the material and lighting, because it represents the distant environmental illumination convolved with an object's specular BRDF, hence preventing relighting. In contrast to a global environment map, Azinovic \etal~\cite{azinovic2019inverse} model lighting as surface emissions, and use a Monte Carlo differentiable renderer to jointly estimate material properties and surface emissions from multi-view images conditioned on scanned geometry and object segmentation masks.
Other work seeks to predict 
illumination, materials and shape from a single image via learning-based priors~\cite{barron2014shape,li2020inverse,blind}.  Ramamoorthi and Hanrahan~\cite{ravi_signal} estimates BRDF and lighting via deconvolution given known geometry. In our work, we aim to jointly estimate the material and environment, together with geometry and surface normals, solely from multi-view 2D images under the challenging setting of unknown static natural illumination.

\begin{figure}[!tb]
    \centering
    \includegraphics[width=1.0  \columnwidth]{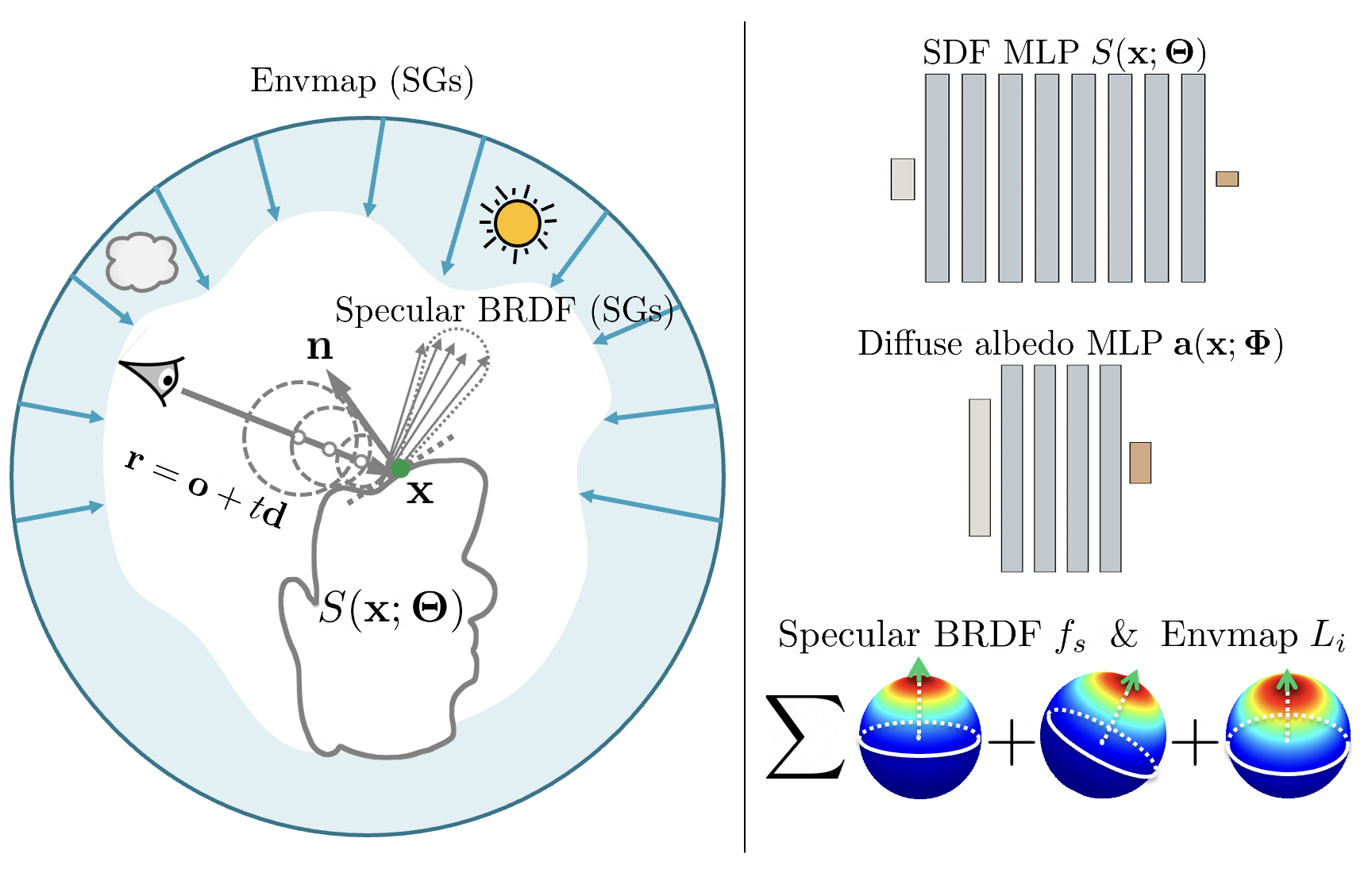}
    \vspace{-1.0\baselineskip}
    \caption{Overview of our PhySG inverse rendering pipeline. To render the color for a camera ray $\mathbf{r}=\mathbf{o}+t\mathbf{d}$, we first use  sphere tracing to find the ray's intersection $\mathbf{x}$ with the geometry in the form of a signed distance function (SDF) represented as an MLP $S(\mathbf{x};\boldsymbol{\Theta})$. The surface normal 
    $\mathbf{n}={\nabla_\mathbf{x} S}$ at location $\mathbf{x}$ is then computed as the SDF gradient. We also represent the spatially-varying diffuse albedo $\mathbf{a}(\mathbf{x};\boldsymbol{\Phi})$ with an MLP.
    Given the surface normal, albedo, and viewing direction at $\mathbf{x}$, we render a color using the spherical Gaussian (SG) renderer, where we represent both the environment map and a specular BRDF using SGs. The rendered image can then be compared to the ground-truth via image reconstruction loss to jointly optimize the unknowns: geometry and surface normal, spatially-varying diffuse albedo, specular BRDF and environment map. 
    }\label{fig:pipeline_overview}
    \vspace{-1.0\baselineskip}
\end{figure}

\medskip
\noindent\textbf{Joint shape and appearance refinement.} Given the initial geometry and appearance from RGBD sensors, Maier et al.~\cite{maier2017intrinsic3d} and Zollhofer et al.~\cite{zollhofer2015shading} jointly refine geometry and appearance by assuming Lambertian BRDF and incorporating shading cues. They pre-compute a lighting model based on spherical harmonics~\cite{ramamoorthi2001efficient}, then fix it while optimizing the shape and diffuse albedo. They adopt voxelized SDFs as their geometric representation.
Assuming known illumination, Oxholm and Nishino~\cite{oxholm2014multiview} also exploit reflectance cues to refine geometry computed via visual hulls. In contrast to these prior works, our method does not require scanned geometry or a known environment map. Instead, we estimate material and lighting parameters, as well as geometry and surface normals in an end-to-end fashion.

\medskip
\noindent\textbf{The rendering equation. } Kajiya et al.~\cite{kajiya1986rendering} proposed the rendering equation based on the physical law of energy conservation. For a surface point $\location$ with surface normal $\normaldir$, suppose $ L_i(\lightdir;\location)$ is the incident light intensity at location $\location$ along the direction $\lightdir$, and BRDF $f_r(\viewdir, \lightdir; \location)$ is the reflectance coefficient of the material at location $\location$ for incident light direction $\lightdir$ and viewing direction $\viewdir$, then the observed light intensity $L_o(\viewdir; \location)$ is an integral over the hemisphere $\Omega=\{\lightdir:\lightdir\cdot \normaldir>0\}$~\footnote{Viewing direction $\viewdir$, lighting direction $\lightdir$ and surface normal $\normaldir$ are all assumed to point away from the scene.}:
\begin{equation}
\begin{aligned}
\label{eq:rendering_eq}
    L_o(\viewdir; \location)=\int_{\Omega}  L_i(\lightdir)\, f_r(\viewdir, \lightdir; \location)\, (\lightdir \cdot \normaldir)  \D \lightdir. 
\end{aligned}
\end{equation}
The BRDF $f_r(\viewdir, \lightdir; \location)$ is a function of viewing direction $\viewdir$, and models view-dependent effects such as specularity.

%% file: sections/03-method.tex
\section{Method}
In this section, we describe our PhySG pipeline and its three major components: (1) geometry modeling, (2) appearance modeling, and (3) forward rendering. These components are designed to be differentiable, so that the whole pipeline can be optimized end-to-end from multiple images captured under static illumination.

\medskip
\noindent \textbf{Geometry modeling.} Motivated by the success of signed distance functions (SDFs) for representing shape~\cite{yariv2020multiview,jiang2020sdfdiff,Park_2019_CVPR,maier2017intrinsic3d,zollhofer2015shading}, 
we adopt SDFs 
as our 
geometric representation. SDFs support ray casting via sphere tracing, are differentiable, and automatically satisfy the constraint between shape and surface normal---the surface normal is exactly the gradient of the SDF.
We represent SDFs with MLPs (rather than voxel grids) for their memory efficiency and infinite resolution~\cite{Park_2019_CVPR}. 
Concretely, let $S(\mathbf{x};\boldsymbol{\Theta})$ be our SDF,\footnote{We assume SDF$>$0 is an object's exterior, while SDF$<$0 is its interior.} where $\mathbf{x}$ is a 3D point and $\boldsymbol{\Theta}$ are the MLP weights. Our MLP consists of 8 nonlinear layers of width 512, with a skip connection at $4^{th}$ layer. To allow the MLP to model high-frequency geometric detail, we use positional encoding  with 6 frequency components to encode the location of a 3D point~\cite{tancik2020fourier,mildenhall2020nerf}.\footnote{Using $L$ frequency components, positional encoding maps vector $\mathbf{p}$ to $\big(\mathbf{p}, \sin(2^0\mathbf{p}), \cos(2^0\mathbf{p}),\dots, \sin(2^{L-1}\mathbf{p}), \cos(2^{L-1}\mathbf{p})\big)$.}
An alternate to SDFs is to use occupancy fields~\cite{mescheder2019occupancy,niemeyer2020differentiable}, 
but ray tracing through occupancy fields is much slower, requiring 
root-finding to locate the surface. 
While occupancy fields require over 100 MLP evaluations per cast ray~\cite{niemeyer2020differentiable},
for SDFs, the MLP only needs to be evaluated $\sim$10 times via sphere tracing.

To render the pixel color for a camera ray, we first find the ray's point of intersection with the SDF 
by starting from the ray's intersection with the object bounding box and marching along the ray via sphere tracing as in~\cite{yariv2020multiview}, where the size of each step is the signed distance at the current location. The intersection point's location $\mathbf{x}$ and surface normal 
$\mathbf{n}={\nabla_{\mathbf{x}}S}$
are then used by our appearance component to render the pixel's color. Hence, to optimize the geometry, gradients must 
back-propagate through both $\mathbf{x}$  and $\mathbf{n}$ to the SDF parameters $\boldsymbol{\Theta}$. 
Back-propagating through the surface normal $\mathbf{n}$ is straightforward 
via auto-differentiation~\cite{pytorch2019_9015}. To back-propagate through the surface location $\mathbf{x}$, we use the implicit differentiation method presented in~\cite{niemeyer2020differentiable,yariv2020multiview}.
Note however that the sphere tracing algorithm itself need not be differentiable, hence it is very memory-efficient. 

\medskip
\noindent \textbf{Appearance modeling.} To model
a single-material specular object in a way consistent with the rendering equation (see Eq.~\ref{eq:rendering_eq}), we use two optimizable components: (1) an environment map, and (2) BRDF consisting of spatially varying diffuse albedo and a shared monochrome isotropic specular component. Note however that we do not model self-occlusion or indirect illumination.
The hemispherical integral in the rendering equation generally does not have a closed-form expression, necessitating 
expensive Monte-Carlo methods for numeric evaluation. However, in our setting of glossy material and distant direct illumination, we can utilize spherical Gaussians (SGs)~\cite{yan2012accurate} to efficiently approximate
the rendering equation in closed form. 

An $n$-dimensional spherical Gaussian (SG) is a spherical function that takes the form~\cite{wang2009all}:
\begin{equation}\label{eq:sg} 
    G(\boldsymbol{\nu}; \lightsgdir, \lightsgsharp, \lightsgamp) = \lightsgamp \, e^{\lightsgsharp (\boldsymbol{\nu} \cdot \lightsgdir - 1)},
\end{equation}
where $\boldsymbol{\nu} \in \mathbb{S}^2$ is the function input, $\lightsgdir \in \mathbb{S}^2$ is the lobe axis, $\lightsgsharp \in \mathbb{R}_+$ is the lobe sharpness, and $\lightsgamp \in \mathbb{R}_+^n$ is the lobe amplitude.
Our environment map $L_i(\lightdir;\location)=L(\lightdir)$\footnote{We drop the location $\mathbf{x}$ due to the distant illumination assumption.} is then represented with a mixture of $M=128$ SGs:
\begin{align}
    L_i(\lightdir)=\sum_{k=1}^M G(\lightdir;\lightsgdir_k, \lightsgsharp_k, \lightsgamp_k).
\end{align}
We represent the spatially-varying diffuse albedo with an MLP mapping a surface point $\mathbf{x}$ to a color vector $\mathbf{a}$, i.e., $\mathbf{a}(\mathbf{x;\boldsymbol{\Phi}})$. Positional encoding is also applied to fit high-frequency texture details~\cite{tancik2020fourier,martin2020nerf}. Specifically, we use an MLP with 4 nonlinear layers of width 512, and encode location $\mathbf{x}$ with $10$ frequencies. As for the shared specular component, we use the same simplified Disney BRDF model~\cite{burley2012physically,karis2013real} as in prior work~\cite{bi2020deep,li2018learning}:  
\begin{equation}\label{eq:microfacet}
\begin{aligned}
    f_s(\viewdir, \lightdir) &=\mathcal{M}(\viewdir, \lightdir) \, \mathcal{D}(\mathbf{h}),
\end{aligned}
\end{equation}
where $\mathbf{h} = (\viewdir + \lightdir) / \|\viewdir + \lightdir\|_2$, $\mathcal{M}$ accounts for the Fresnel and shadowing effects, and $\mathcal{D}$ is the normalized distribution function. We include details of $\mathcal{M}$ and $\mathcal{D}$ in the supplemental material. We represent $\mathcal{D}$ with a single SG:
\begin{align}
    \mathcal{D}(\mathbf{h})=G(\mathbf{h}; 
\brdfsgdir,\brdfsgsharp,\brdfsgamp).
\end{align}
Our isotropic specular BRDF assumption results in $\brdfsgdir$ aligning with surface normal, i.e., $\brdfsgdir=\normaldir$, while the monochrome assumption makes the three numbers in $\brdfsgamp$ identical.

To evaluate the rendering equation at a point $\mathbf{x}$  with surface normal $\mathbf{n}$ viewed along direction $\viewdir$,  $\mathcal{D}$ must be spherically warped, while $\mathcal{M}$ must be approximated by a constant at this specific location $\mathbf{x}$~\cite{wang2009all}:
\begin{align}
\mathcal{D}_\location(\mathbf{h})&=G(\mathbf{h}; 
\normaldir,\frac{\brdfsgsharp}{4\mathbf{h}\cdot \viewdir}, \brdfsgamp),\\
\mathcal{M}_\location(\viewdir, \lightdir)&\approx \mathcal{M}(\viewdir, 2(\viewdir\cdot\normaldir)\normaldir-\viewdir).
\end{align}
Hence for the point $\mathbf{x}$, we have:
\begin{align}
f_s(\viewdir, \lightdir; \location)=G(\mathbf{h}; \normaldir, \frac{\brdfsgsharp}{4\mathbf{h}\cdot \viewdir}, 
\mathcal{M}_{\mathbf{x}}\brdfsgamp).
\end{align}
Now that both $L_i(\lightdir)$ and $f_r(\viewdir,\lightdir;\location)\hspace{-1mm}=\hspace{-1mm}\frac{\mathbf{a}}{\pi}+f_s(\viewdir, \lightdir; \location)$ in the rendering equation are represented with SGs, 
we further approximate the remaining term $\lightdir\cdot \mathbf{n}$ with a SG~\cite{hemisphere_int}:
\begin{align}
\lightdir\cdot \mathbf{n}\approx G(\lightdir;0.0315,\mathbf{n}, 32.7080)-31.7003.
\end{align}
Finally, we integrate the multiplication of these SGs in closed-form~\cite{hemisphere_int} to compute the observed color~$L_o(\viewdir;\location)$.

To summarize, the optimizable parameters in our appearance component are $\big\{\lightsgdir_k,\lightsgsharp_k, \lightsgamp_k\big\}_{k=1}^M$, $\{\brdfsgsharp, \brdfsgamp\}$, and $\boldsymbol{\Phi}$, which are parameters of the environment map, specular BRDF, and spatially-varying diffuse albedo, respectively. 

\medskip
\noindent \textbf{Forward rendering.} Given our geometric and appearance components, we perform forward rendering of a ray's color as follows: (1) use sphere tracing to find the intersection point $\mathbf{x}$ between the ray $\mathbf{r}=\mathbf{o}+t\mathbf{d}$ and the surface
$S(\mathbf{x};\boldsymbol{\Theta})$;
\ \ (2) compute the surface normal $\mathbf{n}=\nabla_{\mathbf{x}}S$ at 
$\mathbf{x}$ via automatic differentiation~\cite{pytorch2019_9015}; \ \ (3) compute the diffuse albedo $\mathbf{a}(\mathbf{x;\boldsymbol{\Phi}})$ at $\mathbf{x}$; \ \ (4) use the surface normal $\mathbf{n}$, environment map  $\big\{\lightsgdir_k, \lightsgsharp_k, \lightsgamp_k\big\}_{k=1}^M$,  diffuse albedo $\mathbf{a}$, specular BRDF $\{\brdfsgsharp,\brdfsgamp\}$, and viewing direction $\mathbf{d}$, to compute the color for ray $\mathbf{r}$ by evaluating the rendering equation in closed form with our SG approximation. This procedure is illustrated in Fig.~\ref{fig:pipeline_overview}.

We now show that our pipeline is fully differentiable, in that its output (the rendered color) is differentiable w.r.t.\ all the optimizable parameters. First, the rendered color is differentiable w.r.t.\ the variables $\mathbf{n}, \big\{\lightsgdir_k, \lightsgsharp_k, \lightsgamp_k\big\}_{k=1}^M,\mathbf{a},\{\brdfsgsharp,\brdfsgamp\}$ in step (4), because the SG renderer is simply the closed-form integration of spherical Gaussians. Since  the diffuse albedo $\mathbf{a}=\mathbf{a}(\mathbf{x};\boldsymbol{\Phi})$ is an MLP in step (3), the rendered color is differentiable w.r.t.\ $\mathbf{x}$ and $\boldsymbol{\Phi}$ by the chain rule. For our geometric model, we have shown that there exist gradients of both the surface location $\mathbf{x}$ and surface normal $\mathbf{n}$ w.r.t.\ the SDF parameters $\boldsymbol{\Theta}$. Thus by the chain rule, the rendered color is differentiable w.r.t.\ $\boldsymbol{\Theta}$ as well.

\medskip
\noindent \textbf{Loss functions.} To optimize parameters given a set of images,
we render images from the same viewpoints as the input images, and compute an $\ell_1$ image reconstruction loss. We also enforce non-negative minimum SDF values along non-object pixel rays indicated by object segmentation masks, and regularize the SDF's gradient to have unit norm~\cite{gropp2020implicit}. Concretely, at each training iteration, we first randomly sample a batch of pixels consisting of: object pixels $\mathbf{r}^{\mathit{obj}}_i$ with ground-truth color $\big\{\mathbf{c}^\mathit{gt}_i\big\}_{i=1}^{N_\mathit{obj}}$, and non-object pixels $\big\{\mathbf{r}^\mathit{nobj}_i\big\}_{i=1}^{N_\mathit{nobj}}$. Then we render colors $\mathbf{c}_i^\mathit{obj}$ for $\mathbf{r}^\mathit{obj}_i$, while finding the minimal SDF value $S^\mathit{nobj}_i$ along camera rays $\mathbf{r}_i^\mathit{nobj}$ by taking the minimal SDF value among 100 points uniformly lying on the ray segment inside object bounding box. We also randomly sample $\big\{\mathbf{x}_i\big\}_{i=1}^{N_\mathit{x}}$ inside the object bounding box. Our full loss is:
\begin{align}\label{eq:loss_fn}
\ell = &\frac{1}{N_\mathit{obj}}\sum_{i=1}^{N_\mathit{obj}}\big\Vert \mathbf{c}^\mathit{obj}_i-\mathbf{c}^{\mathit{gt}}_i\big\Vert_1
\nonumber \\
&+\beta_1 \frac{1}{N_\mathit{nobj}}\sum_{i=1}^{N_\mathit{nobj}}\frac{\ln(1+e^{-{\alpha S_i^{\mathit{nobj}}}})}{\alpha}
\nonumber\\
&+\beta_2 \frac{1}{N_\mathit{x}}\sum_{i=1}^{N_\mathit{x}}\big\Vert \Vert\nabla_{\mathbf{x}_i}{S}\Vert_2 - 1\big\Vert_2^2,
\end{align}
where $\frac{\ln(1+e^{-{\alpha S_i^{\mathit{nobj}}}})}{\alpha},\alpha\hspace{-1mm}>\hspace{-1mm}0$  is a smooth approximation of a horizontally flipped ReLU $\max\{-S_i^\mathit{nobj},0\}$ (larger $\alpha$ yields tighter approximation); and $\beta_1$ and $\beta_2$ are weights balancing different loss terms. We set $\beta_1\hspace{-1mm}=\hspace{-1mm}100,\beta_2\hspace{-1mm}=\hspace{-1mm}0.1,N_{\mathit{obj}}\hspace{-1mm}+\hspace{-1mm}N_{\mathit{nobj}}\hspace{-1mm}=\hspace{-1mm}2048,N_{\mathit{x}}\hspace{-1mm}=\hspace{-1mm}1024$ in our experiments; $\alpha$ gradually grows from 50 to 1600 as suggested in~\cite{yariv2020multiview}. Finally, rather than sampling $N_{\mathit{obj}} + N_{\mathit{nobj}}$  independent pixels, we sample $\frac{N_{\mathit{obj}}+N_{\mathit{nobj}}}{4}$ patches of size $2 \times 2$, and add an additional loss term to penalize the variance of surface normals inside patches consisting only of object pixels. We set the weight for this smoothness loss to 10. We train on a single 12GB NVIDIA GPU for 250k iterations.

\medskip
\noindent\textbf{Initialization.} The SDF weights $\boldsymbol{\Theta}$ 
are initialized using the method of~\cite{gropp2020implicit} such that the initial shape is roughly a sphere. The diffuse albedo $\mathbf{a}(\mathbf{x};\boldsymbol{\Phi})$ is initialized such that predicted albedo is $\sim$0.5 at all locations inside the object bounding box. For the specular BRDF, the initial lobe sharpness $\brdfsgsharp$ is randomly drawn from $[95, 125]$, while the initial specular albedo $\brdfsgamp$ is randomly drawn from $[0.18, 0.26]$. For the environment map, the lobes are initialized to distribute uniformly on the unit sphere using a spherical Fibonacci lattice~\cite{keinert2015spherical}, with monochrome colors; we also scale the randomly initialized lobes' amplitude so that the initial rendered pixel intensity output by our pipeline is $\sim$0.5. In addition, since different captures can vary significantly in exposure, we scale all input images of an object with the same constant such that the median intensity of all scaled images is 0.5. We empirically find that if the initial environment map is too bright or too dark, the diffuse albedo MLP sometimes gets stuck, predicting all zeros or ones during training. Our proposed initialization addresses this issue.

%% file: sections/04-experiment.tex
\section{Experiments}
We perform experiments on both synthetic and real-world data to validate our PhySG pipeline. 

\subsection{Synthetic data}
\begin{figure}[htb]
\centering
    \begin{tabular}{c@{\hspace{0.02em}}c@{\hspace{0.02em}}}
        \includegraphics[width=0.48\columnwidth]{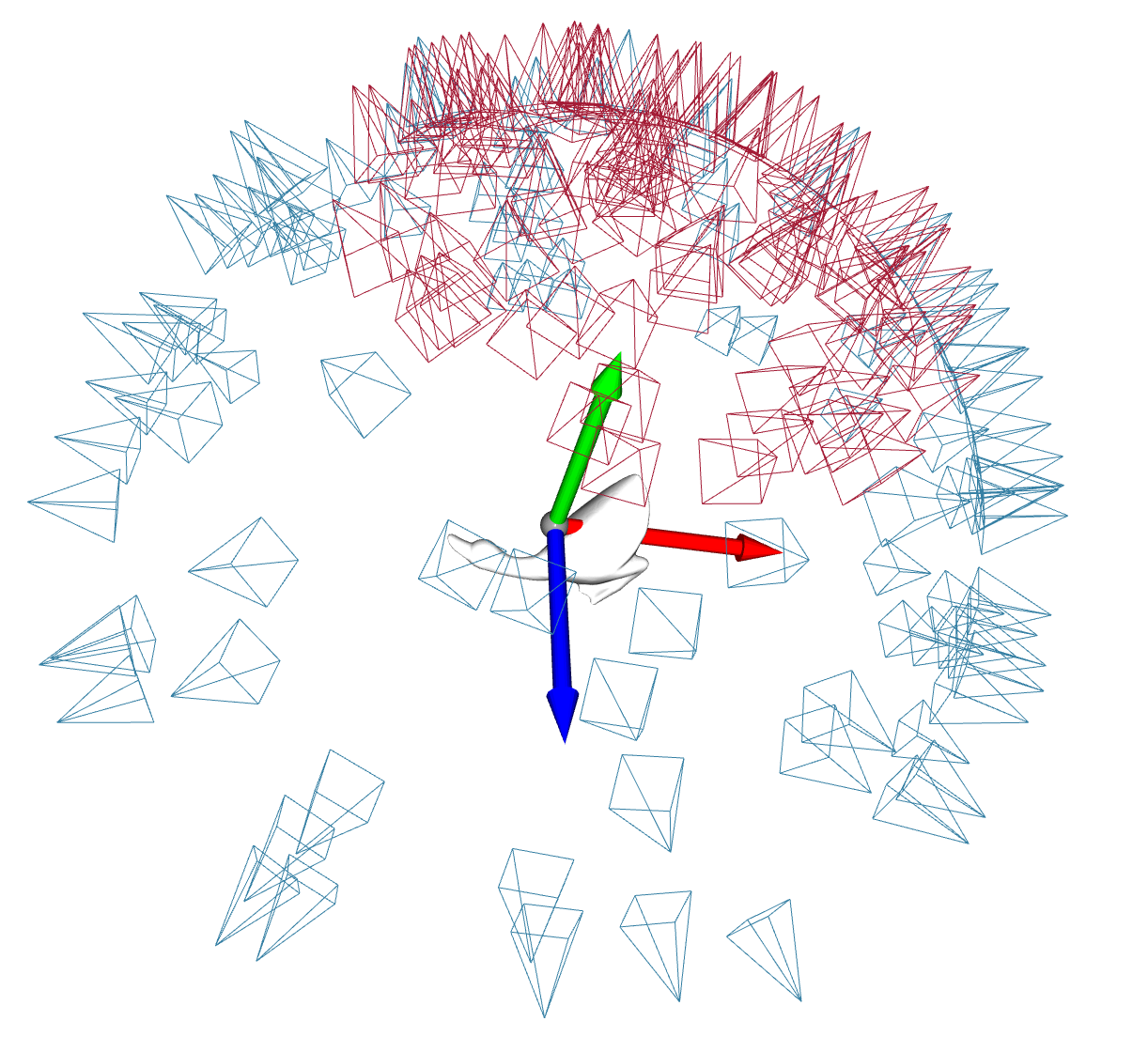}  &
        \includegraphics[width=0.48\columnwidth]{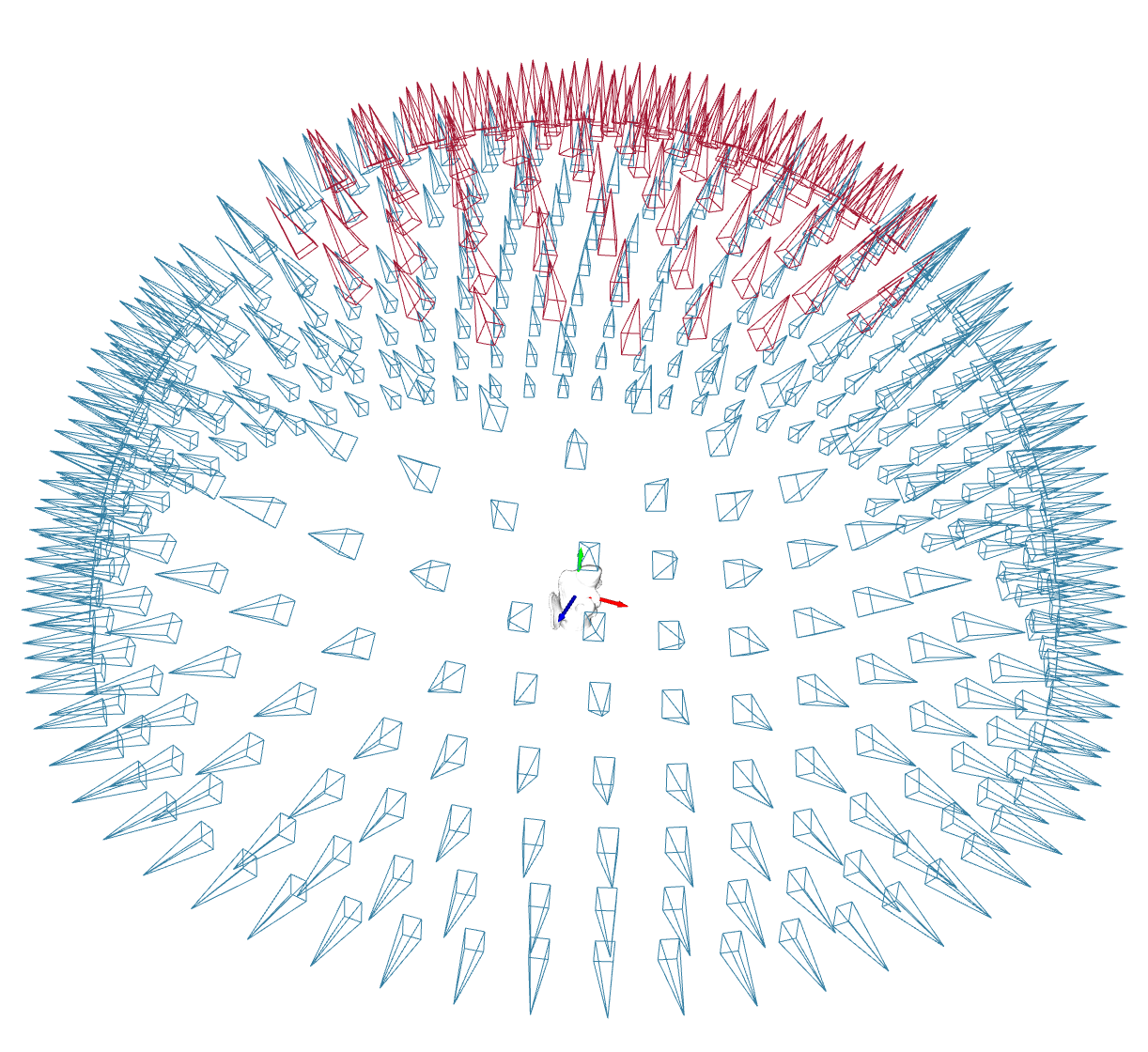}
     \end{tabular} 
  \vspace{-1.0\baselineskip}
  \caption{Camera setup for synthetic duck (left)~\cite{dong2014appearance} and real SLF fish (right)~\cite{wood2000surface}. Blue for training, red for testing. Note that testing viewpoints deviate significantly from training ones. Latitudes of testing cameras are above 55 degrees, while training cameras are below this latitude threshold. 
  }\label{fig:camera_setup}
\end{figure}

To create synthetic data, we use objects from~\cite{dong2014appearance,zhou2016sparse-as-possible}; for each object, we render 200 images with colored environmental lighting using the Mitsuba renderer~\cite{Mitsuba}, 100 each for training and testing. To test the extrapolation capability of different algorithms, the test images are distributed inside a 70-degree cone around the object's north pole, while the training images cover the rest of the upper hemisphere (see Fig.~\ref{fig:camera_setup}). We use the Ward BRDF model~\cite{ward1992measuring} included in Mitsuba, and set the specular albedo to $(0.3, 0.3, 0.3)$ and roughness values along the tangent and bitangent directions to $0.05$. Ground truth surface normal maps and diffuse albedos are also rendered to quantitatively evaluate our inverse rendering results. To evaluate the relighting performance of our pipeline, we also render the same object with two other environment maps in Mitsuba to serve as ground truth.

We report image quality metrics: LPIPS~\cite{zhang2018perceptual}, SSIM, and PSNR on held-out test viewpoints. As there is an inherent scale ambiguity in inverse rendering problems, before computing the metrics, we align our predicted image $\hat{I}$ to ground-truth $I$ via channel-wise scale factors. Specifically, let $\hat{I}_r,I_r$ denote the red channel of $\hat{I},I$, respectively. Then the scale factor $s_r$ for the red channel is estimated via:
\begin{align} \label{eq:align_scale}
    s_r=\text{Median}({I_r}\big/{\hat{I}_r}).
\end{align}
The green and blue channels are scaled similarly.

As shown by the quantitative evaluation in Tab.~\ref{tab:syn_inverse} and qualitative evaluation in Fig.~\ref{fig:main_syn}, our synthesized novel test views, estimated diffuse albedo and surface normal, as well as material editing and relighting results closely match the ground truth on synthetic data, despite the test viewpoints representing a difficult view extrapolation scenario. Note especially that our method correctly extrapolates the challenging specular highlight in Fig.~\ref{fig:main_syn}.

\begin{table}
\centering
\begin{tabular}{lccc}
\toprule 
& $\downarrow$LPIPS  & $\uparrow$SSIM  &$\uparrow$PSNR \\
Diffuse albedo & 0.0339 & 0.989& 33.43\\
Novel view  & 0.0170  & 0.990  & 35.93 \\
Relighting & 0.0227 & 0.988 & 33.25\\
\midrule
Surface normal error ($^\circ$)  & \multicolumn{3}{c}{2.528} \\
\bottomrule
\end{tabular}
\caption{Quantitative evaluation of our inverse rendering results on the synthetic dataset. We compare predictions of our pipeline against the ground-truth rendered with Mitsuba. Since there is a scale ambiguity in inverse rendering problems, we align our predictions to ground-truth before evaluating the metrics (see Eq.~\ref{eq:align_scale} for details). We also report the average angular error of our estimated surface normals. 
}\label{tab:syn_inverse}
\end{table}

\begin{table}
\centering
\begin{tabular}{lcc}
\toprule 
&  $\downarrow$Surface Normal Error ($^\circ$) & $\downarrow$Chamfer $L_1$\\
Ours & 2.528 & 0.00142\\
NeRF  & 36.05 &  0.01650\\
IDR & \textbf{2.207} & \textbf{0.00136}\\
DVR & 38.90 & 0.13800 \\
\bottomrule
\end{tabular}
\caption{
Evaluation of recovered geometry on synthetic data. We report avg.\ surface normal error on test views and $L_1$ Chamfer (point-to-mesh) distance between estimated and GT meshes (normalized to have a unit bounding box).
}\label{tab:comp_geometry}
\end{table}

\begin{table*}
\centering
\resizebox{\textwidth}{!}{
\begin{tabular}{l@{\hspace{1em}} c@{\hspace{1em}}c@{\hspace{1em}}c@{\hspace{2em}} c@{\hspace{1em}}c@{\hspace{1em}}c@{\hspace{2em}} c@{\hspace{1em}}c@{\hspace{1em}}c@{\hspace{2em}} c@{\hspace{1em}}c@{\hspace{1em}}c@{\hspace{2em}} c@{\hspace{1em}}c@{\hspace{1em}}c@{\hspace{2em}} ccc}
\toprule 
&\multicolumn{3}{c}{Synthetic Kitty} & \multicolumn{3}{c}{Synthetic Bear} & \multicolumn{3}{c}{Synthetic Duck} &  \multicolumn{3}{c}{Synthetic Mouse} & \multicolumn{3}{c}{SLF fish~\cite{wood2000surface}} & \multicolumn{3}{c}{Chips Corncho1~\cite{park2020seeing}} \\
\# train/test (HxW) &\multicolumn{3}{c}{100/100 (512x512)} & \multicolumn{3}{c}{100/100 (512x512)} & \multicolumn{3}{c}{100/100 (512x512)} &  \multicolumn{3}{c}{100/100 (512x512)} & \multicolumn{3}{c}{419/106 (480x640)} & \multicolumn{3}{c}{1661/345 (480x640)} \\
& $\downarrow$LPIPS & $\uparrow$SSIM & $\uparrow$PSNR & $\downarrow$LPIPS & $\uparrow$SSIM & $\uparrow$PSNR & $\downarrow$LPIPS & $\uparrow$SSIM & $\uparrow$PSNR & $\downarrow$LPIPS & $\uparrow$SSIM & $\uparrow$PSNR & $\downarrow$LPIPS & $\uparrow$SSIM & $\uparrow$PSNR & $\downarrow$LPIPS & $\uparrow$SSIM & $\uparrow$PSNR \\
Ours & \textbf{0.0189} & \textbf{0.989} & \textbf{36.45}   & 0.0200 & \textbf{0.987} & 33.76  & \textbf{0.0081} & \textbf{0.994} & \textbf{38.70} &   \textbf{0.0209} & \textbf{0.987} & \textbf{34.81} & 0.0142 & 0.969 & 30.27 & 0.0477 & 0.969 & 27.44\\
NeRF~\cite{mildenhall2020nerf}  & 0.0534 & 0.971 &  30.75  & 0.0493 &  0.964 &  28.17 & 0.0338 & 0.976 & 29.72 & 0.0772 & 0.948 & 26.30 & 0.0255 &  0.966 &  28.90 & 0.0478 & 0.969 & 27.64\\
IDR~\cite{yariv2020multiview}  & 0.0202 & 0.987 & 35.21   & \textbf{0.0169} & 0.986 & \textbf{33.88}  & 0.0121 & 0.991 & 36.24 & 0.0259 & 0.983 & 32.67 & \textbf{0.0129} &   \textbf{0.977}  & \textbf{31.48} & \textbf{0.0414} & \textbf{0.975} & \textbf{27.92}\\
DVR~\cite{niemeyer2020differentiable}  & 0.132 & 0.926 & 24.69    & 0.124 & 0.911 &  19.24 & 0.100 & 0.944 & 25.78 & 0.165 & 0.903 & 22.37 & 0.0494 & 0.952 & 22.90 & 0.255 & 0.848 & 16.06\\
\bottomrule
\end{tabular}
}
\caption{We compare the novel test view quality of our method with that of NeRF~\cite{mildenhall2020nerf}, IDR~\cite{yariv2020multiview} and DVR~\cite{niemeyer2020differentiable}.  For synthetic data, HDR images are tonemapped with $I_{\mathit{out}}=I_{\mathit{in}}^{1/2.2}$ and clipped to $[0, 1]$ before computing the metrics. Note that the baseline methods model appearance as surface light field~\cite{wood2000surface}, hence they can not do editing/relighting like ours. On real-world data, our metric numbers are slightly worse than IDR --- this is likely caused by the bias in our physics-based appearance modeling that does not align perfectly with real material properties, while surface light field modeling has little bias but precludes material editing and relighting. 
}\label{tab:compare_novelview}
\end{table*}

\begin{figure*}[htb]
\centering
\begin{tabular}{c@{\hspace{0.02em}}c@{\hspace{0.02em}}c@{\hspace{0.02em}}c@{\hspace{0.02em}}c@{\hspace{0.02em}}}
    {\small GT} & {\small Ours} & {\small NeRF} & {\small IDR} & {\small DVR}
    \\
    \includegraphics[trim=0 0 0 0,clip,width=0.19\textwidth,angle=0]{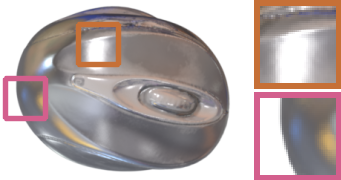}
    &   
    \includegraphics[trim=0 0 0 0,clip,width=0.19\textwidth,angle=0]{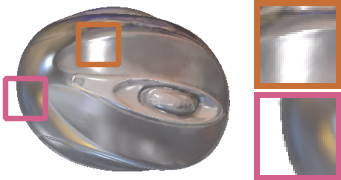}
    &
    \includegraphics[trim=0 0 0 0,clip,width=0.19\textwidth,angle=0]{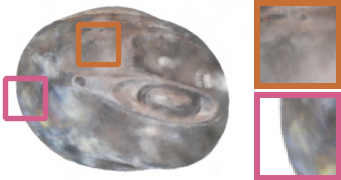} 
    &
    \includegraphics[trim=0 0 0 0,clip,width=0.19\textwidth,angle=0]{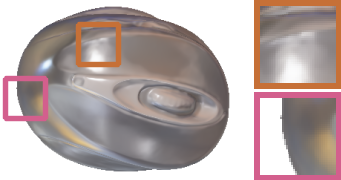}  
    &
    \includegraphics[trim=0 0 0 0,clip,width=0.19\textwidth,angle=0]{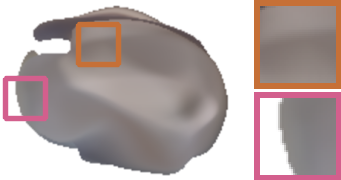}  
    \\
    \includegraphics[trim=0 0 0 0,clip,width=0.19\textwidth,angle=0]{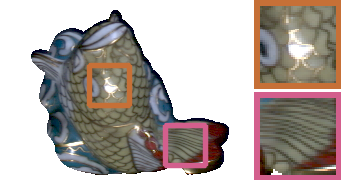}
    &   
    \includegraphics[trim=0 0 0 0,clip,width=0.19\textwidth,angle=0]{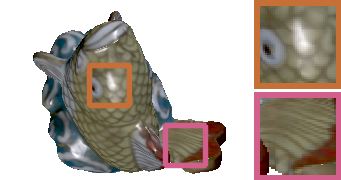}
    &
    \includegraphics[trim=0 0 0 0,clip,width=0.19\textwidth,angle=0]{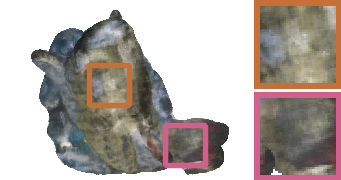} 
    &
    \includegraphics[trim=0 0 0 0,clip,width=0.19\textwidth,angle=0]{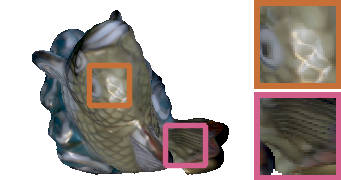}  
    &
    \includegraphics[trim=0 0 0 0,clip,width=0.19\textwidth,angle=0]{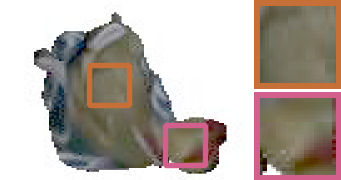}  
    \\
    \includegraphics[trim=0 0 0 0,clip,width=0.19\textwidth,angle=0]{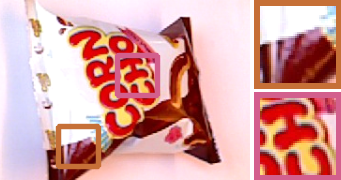}
    &   
    \includegraphics[trim=0 0 0 0,clip,width=0.19\textwidth,angle=0]{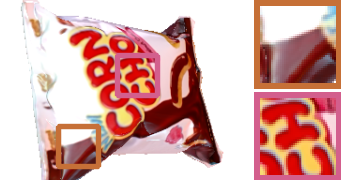}
    &
    \includegraphics[trim=0 0 0 0,clip,width=0.19\textwidth,angle=0]{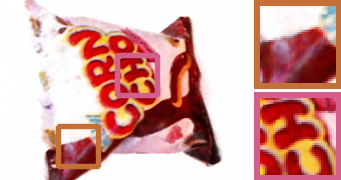} 
    &
    \includegraphics[trim=0 0 0 0,clip,width=0.19\textwidth,angle=0]{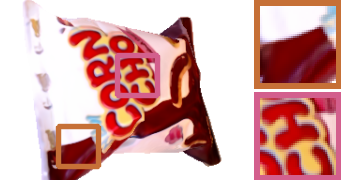}  
    &
    \includegraphics[trim=0 0 0 0,clip,width=0.19\textwidth,angle=0]{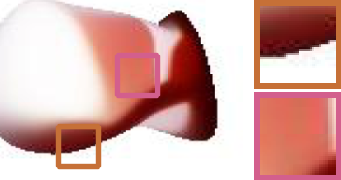}  
    \\
 \end{tabular}  
  \vspace{-0.8\baselineskip}
\caption{On synthetic and real data, we qualitatively compare our novel view extrapolation quality with most related neural rendering techniques:  NeRF~\cite{mildenhall2020nerf}, IDR~\cite{yariv2020multiview} and DVR~\cite{niemeyer2020differentiable}. Our method extrapolates the specularity more reasonably than the baseline methods thanks to our physics-based modeling of the approximate light transport.
}\label{fig:compare_specualr} 
\end{figure*}

\subsection{Real-world data}
We test our method on multiple real-world captures from datasets including SLF~\cite{wood2000surface}, DeepVoxels~\cite{sitzmann2019deepvoxels}, Bag of Chips~\cite{park2020seeing} and DTU~\cite{aanaes2016large}. The objects in these captures are glossy and the illumination is static across different views.

\medskip
\noindent \textbf{SLF dataset.} We use the glossy \emph{fish} from~\cite{wood2000surface}. This dataset is captured with a gantry in a lab-controlled environment. The cameras are distributed on a hemisphere around the center object. We discard images in which the center object has noticeable shadows cast by the gantry or is partly occluded by the platform. Then we split the data according to the cameras' latitudes, with test cameras' latitudes above 55 degrees, as shown in Fig.~\ref{fig:camera_setup}. We render object segmentation masks from the provided laser-scanned meshes.

\input{sections/fig_main_real}

\input{sections/fig_envmap}

\medskip
\noindent \textbf{DeepVoxels.} We use the glossy \emph{globe} and \emph{coffee} objects from DeepVoxels~\cite{sitzmann2019deepvoxels}. These are real-world hand-held captures. The camera parameters are recovered with COLMAP~\cite{schoenberger2016sfm,schoenberger2016mvs}. We use background removal tools~\cite{remove_bg} to automatically generate the object segmentation masks. We leave $\sim$25\% images for testing.

\medskip
\noindent \textbf{Bag of Chips.} We use the glossy \emph{cans} and \emph{corncho1} data from this dataset~\cite{Park_2019_CVPR}. We render object segmentation masks from the provided mesh scanned by RGBD sensors. We leave $\sim$25\% images for testing.

\medskip
\noindent \textbf{DTU dataset.} We use the shiny \emph{scan114 buddha} object from this dataset~\cite{aanaes2016large}. We discard images in which the camera casts noticeable shadows on the object. The object segmentation masks are automatically generated using background removal tools~\cite{remove_bg}. We leave $\sim$25\% images for testing.

\medskip
\noindent Our inverse rendering results are qualitatively shown in Fig.~\ref{fig:main_real}. Video demos are shown in our supplemental material. We can see that our pipeline generates photo-realistic novel views, plausible material editing and relighting results.

\subsection{Comparison with baselines}
We could not identify prior work tackling exactly the same problem as us: simultaneously reconstructing lighting, material, and geometry 
from scratch from 2D images captured under static illumination.
Hence, we compare our PhySG to the most related neural rendering approaches, including NeRF~\cite{martin2020nerf}, IDR~\cite{yariv2020multiview} and DVR~\cite{niemeyer2020differentiable}, in terms of novel view extrapolation quality. 

\input{sections/fig_main_syn}

Like PhySG, these approaches can also be trained end-to-end from 2D image supervision only, but they differ from our method in the way that appearance is modelled. 
Loosely speaking, they all model appearance as an MLP-represented surface light field. In particular,
NeRF maps location $\mathbf{x}$ and viewing direction $\mathbf{d}$ to a color; IDR maps $\mathbf{x}$, $\mathbf{d}$, and surface normal $\mathbf{n}$ to a color, and DVR only takes location $\mathbf{x}$. Tab.~\ref{tab:compare_novelview} and Fig.~\ref{fig:compare_specualr} compares different methods on synthetic and real data. NeRF does poorly in view extrapolation because its volumetric representation does not concentrate colors around surfaces as well as surface-based approaches. Although DVR uses a surface-based shape model, its model does not support view-dependent effects, and so it fails to model 
this kind of glossy data. Compared with DVR, IDR models view-dependence and does a good job in view extrapolation. However, it still has trouble synthesizing specular highlights, due to the lack of a physical model of appearance. In contrast, our method models such highlights well. 
As for geometry, our estimated geometry is nearly as good as IDR’s (and much better than other baselines) shown in Tab.~\ref{tab:comp_geometry}, while we also allow for relighting and material editing.

We also tried redner~\cite{Li:2018:DMC}, a Monte Carlo differentiable renderer representing shapes as meshes. We let redner jointly optimize lighting, texture, BRDF and geometry with an image reconstruction loss. We initialized the mesh a sphere at the beginning of training, and found that redner got stuck in the initial mesh and failed to converge.

\subsection{Robustness to material roughness}
Our method relies on specular highlights to estimate the lighting and material properties. As a result, if the material of interest is purely Lambertian, we face a lighting-texture ambiguity and cannot recover lighting without additional priors. We empirically test the robustness of our pipeline to material roughness on synthetic data. As shown in Fig.~\ref{fig:ablate_roughness}, even from very weak specular highlights, our method can  reconstruct a reasonable-looking environment map.

%% file: sections/fig_main_real.tex
\begin{figure*}[htp]
\centering
    \begin{tabular}{c@{\hspace{0.01em}}c@{\hspace{0.01em}}c@{\hspace{0.01em}}c@{\hspace{0.01em}}c@{\hspace{0.01em}}c@{\hspace{0.01em}}c@{\hspace{0.01em}}}
    {\small Ground-truth} & {\small Ours}  & {\small Diffuse image} & {\small  Edit 1} & {\small Relight 1} & {\small Relight 2} & {\small Esti. Normal}\\
        \includegraphics[trim=0 30 0 35,clip,width=0.1375\textwidth,angle=0]{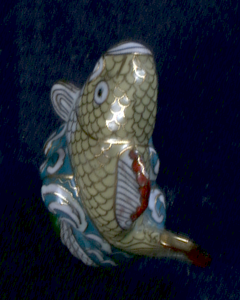}
        &
        \includegraphics[trim=0 30 0 35,clip,width=0.1375\textwidth,angle=0]{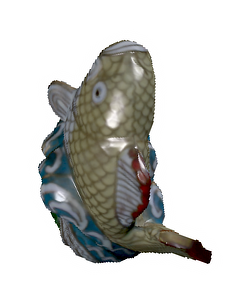}
        &
        \includegraphics[trim=0 30 0 35,clip,width=0.1375\textwidth,angle=0]{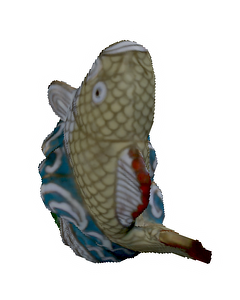}
        &
        \includegraphics[trim=0 30 0 35,clip,width=0.1375\textwidth,angle=0]{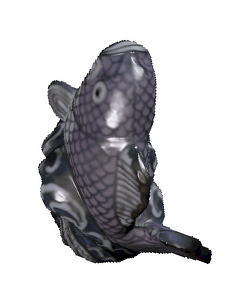} 
        &   
        \includegraphics[trim=0 30 0 35,clip,width=0.1375\textwidth,angle=0]{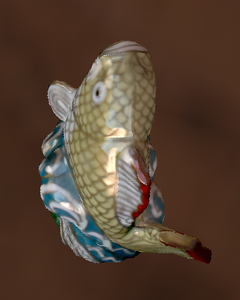}
        &
        \includegraphics[trim=0 30 0 35,clip,width=0.1375\textwidth,angle=0]{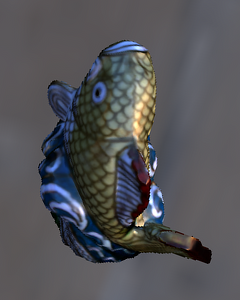}
        &
        \includegraphics[trim=0 30 0 35,clip,width=0.1375\textwidth,angle=0]{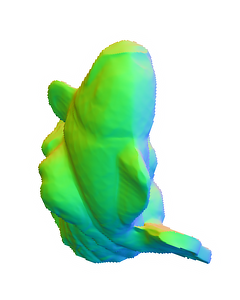}
        \\ 
        \vspace{-1.5em}
        \\
        \includegraphics[trim=205 110 250 0
        ,clip,width=0.1375\textwidth,angle=0]{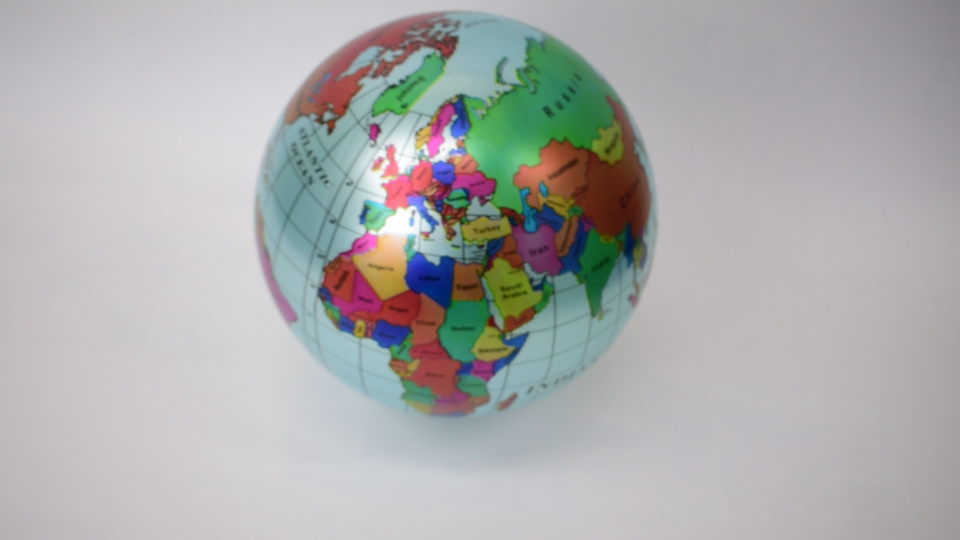}
        &
        \includegraphics[trim=205 110 250 0
        ,clip,width=0.1375\textwidth,angle=0]{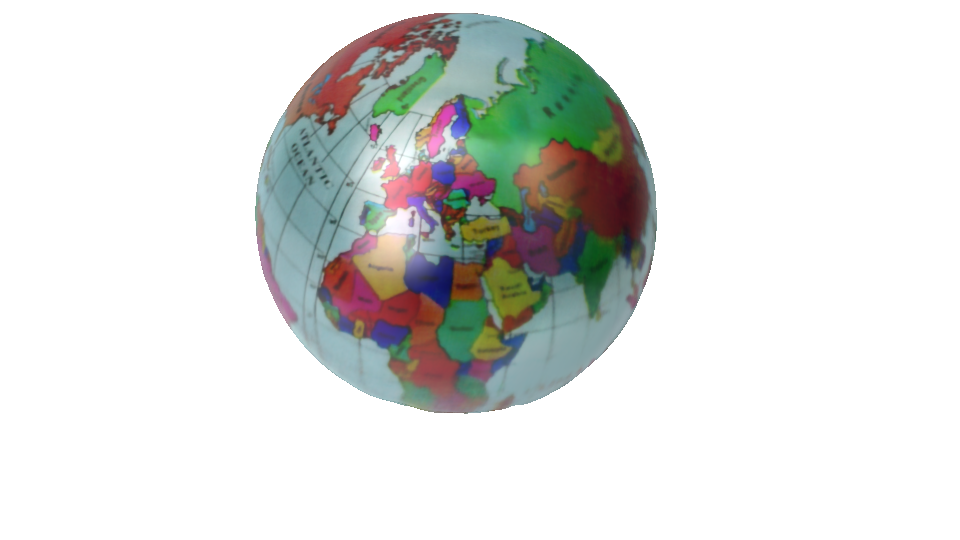}
        &
        \includegraphics[trim=205 110 250 0
        ,clip,width=0.1375\textwidth,angle=0]{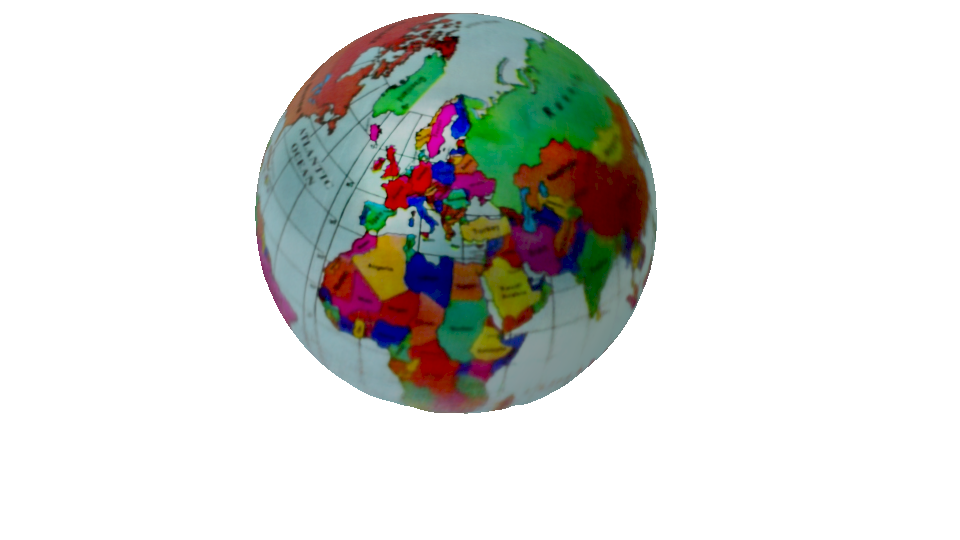}
        &
        \includegraphics[trim=205 110 250 0
        ,clip,width=0.1375\textwidth,angle=0]{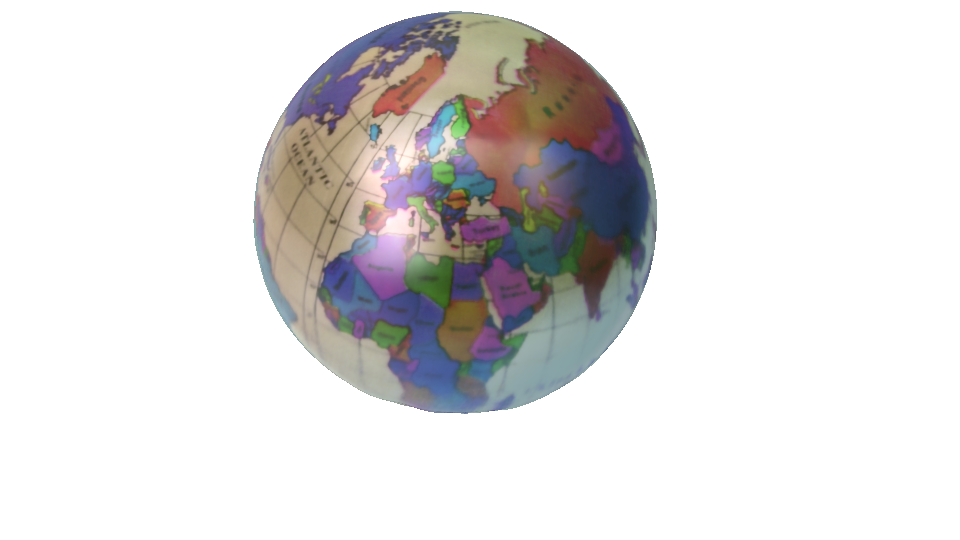}
        &   
        \includegraphics[trim=205 110 250 0
        ,clip,width=0.1375\textwidth,angle=0]{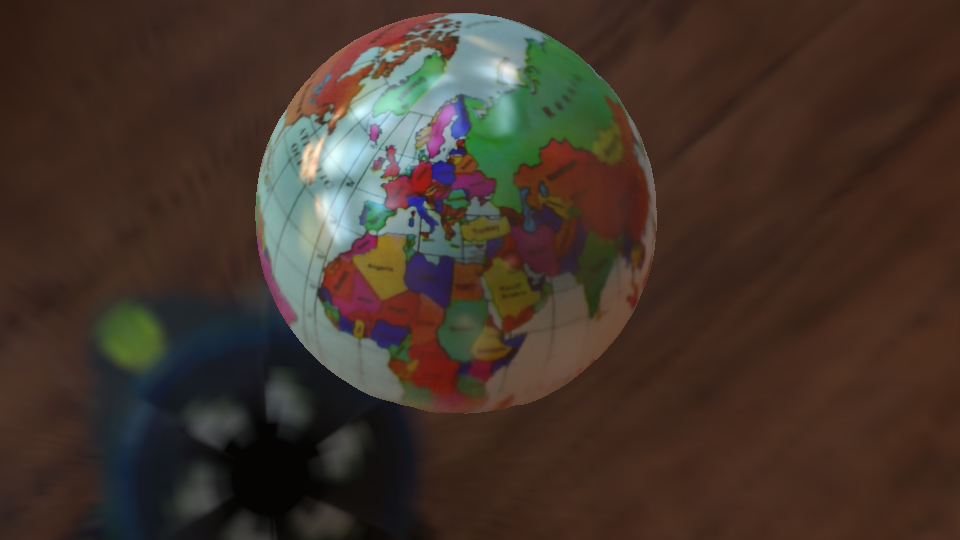}
        &
        \includegraphics[trim=205 110 250 0
        ,clip,width=0.1375\textwidth,angle=0]{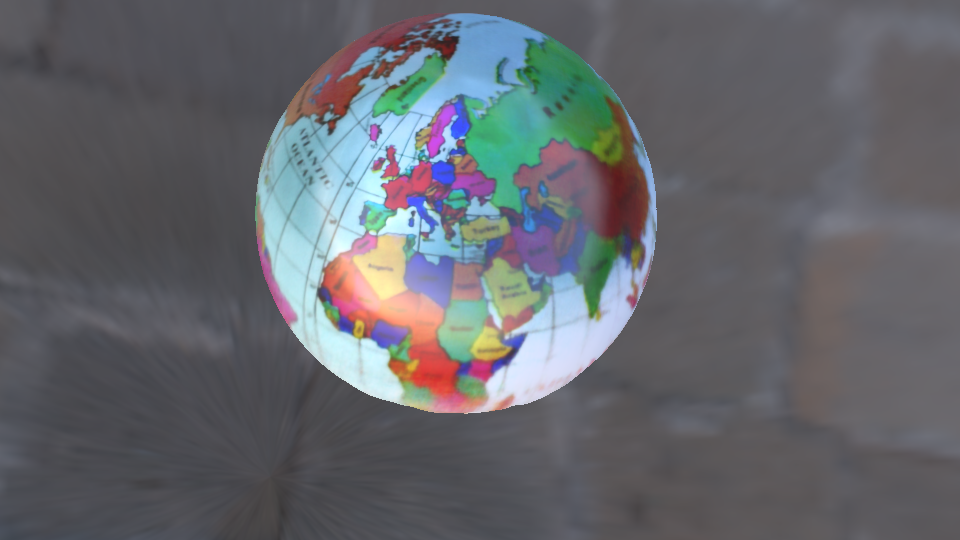}
        &
        \includegraphics[trim=205 110 250 0
        ,clip,width=0.1375\textwidth,angle=0]{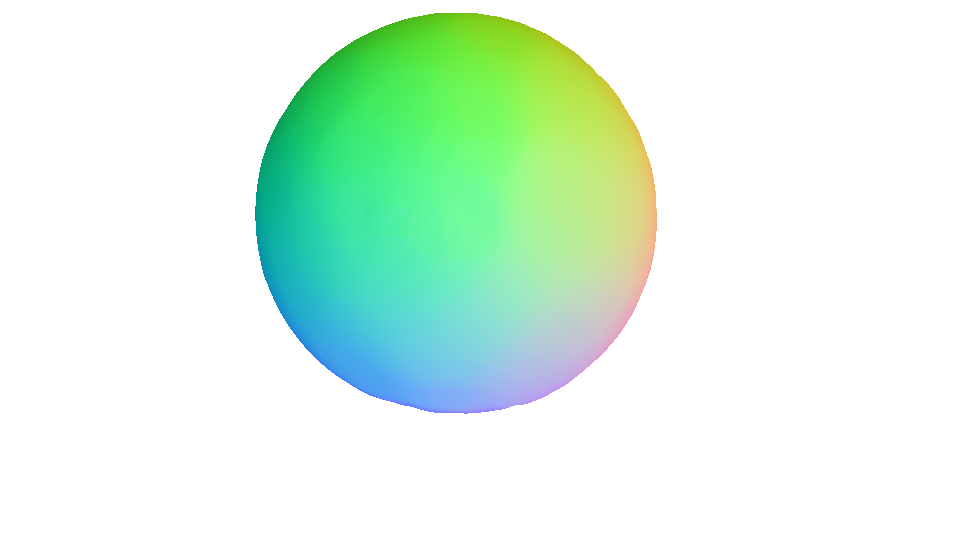}
        \\
        \vspace{-1.5em}
        \\
        \includegraphics[trim=150 0 290 65
        ,clip,width=0.1375\textwidth,angle=0]{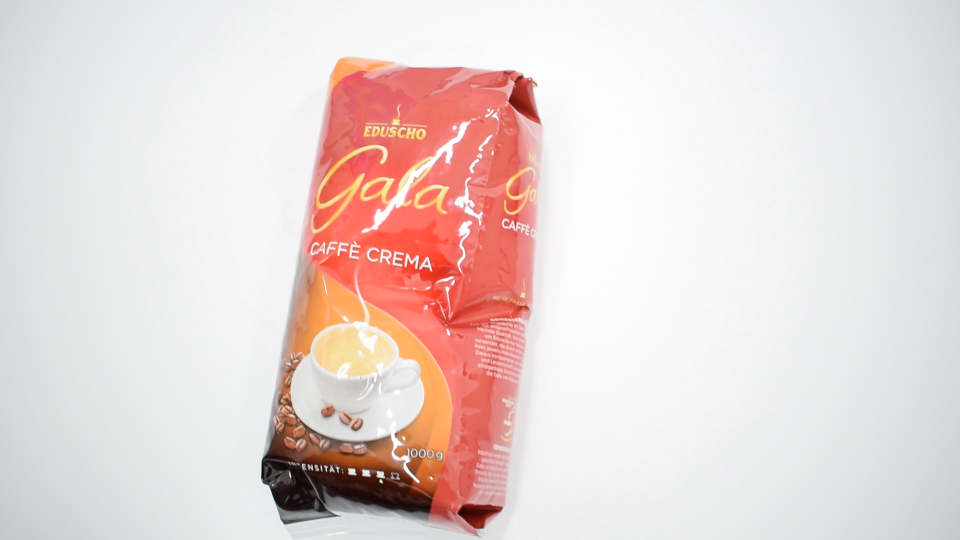}
        &
        \includegraphics[trim=150 0 290 65
        ,clip,width=0.1375\textwidth,angle=0]{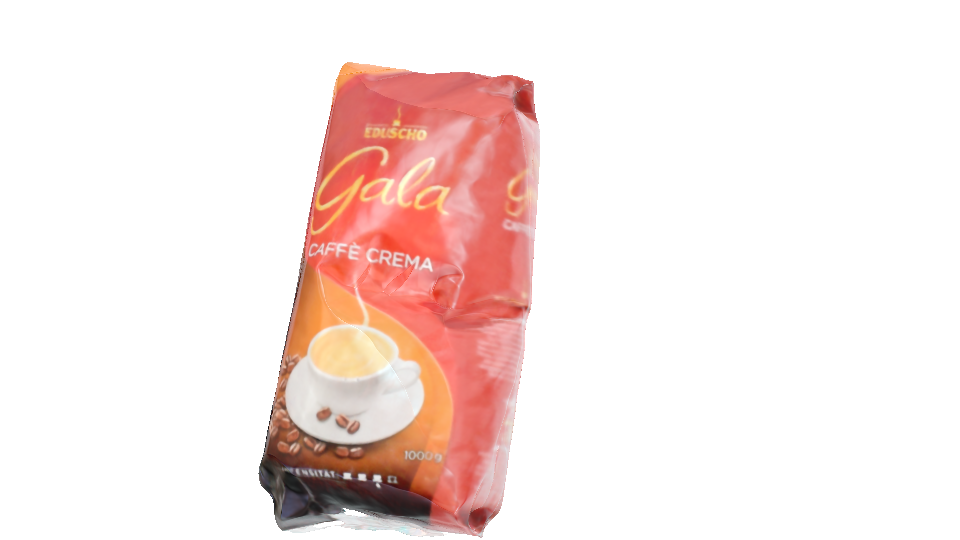}
        &
        \includegraphics[trim=150 0 290 65
        ,clip,width=0.1375\textwidth,angle=0]{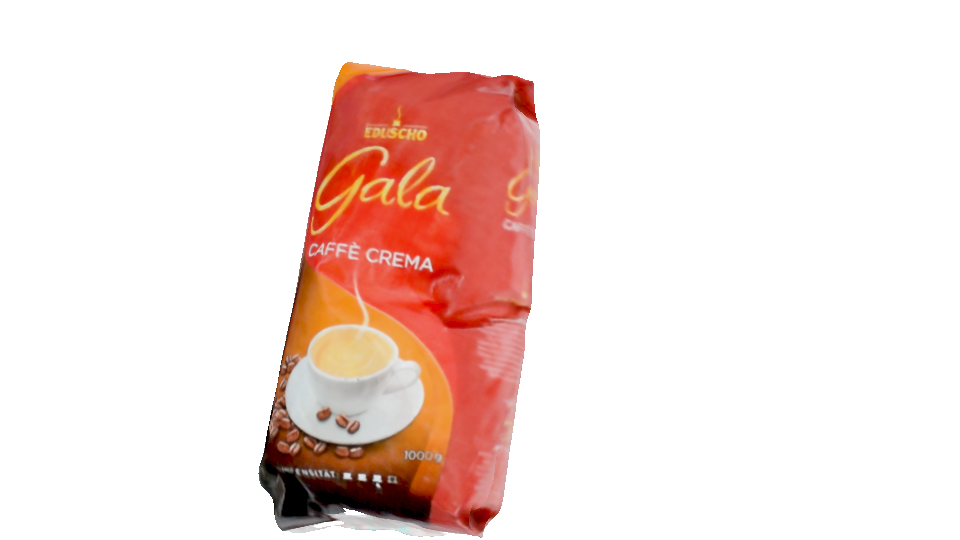}
        &
        \includegraphics[trim=150 0 290 65
        ,clip,width=0.1375\textwidth,angle=0]{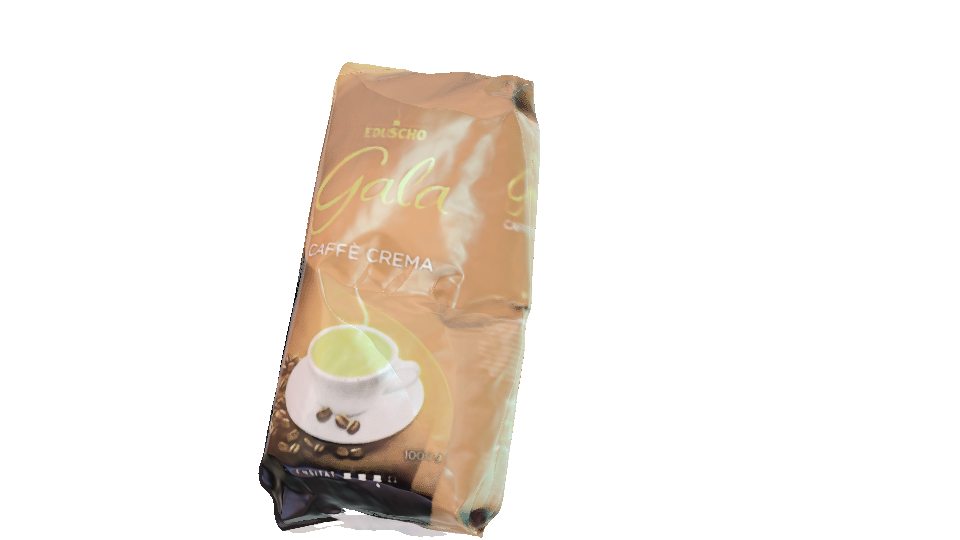}
        &   
        \includegraphics[trim=150 0 290 65
        ,clip,width=0.1375\textwidth,angle=0]{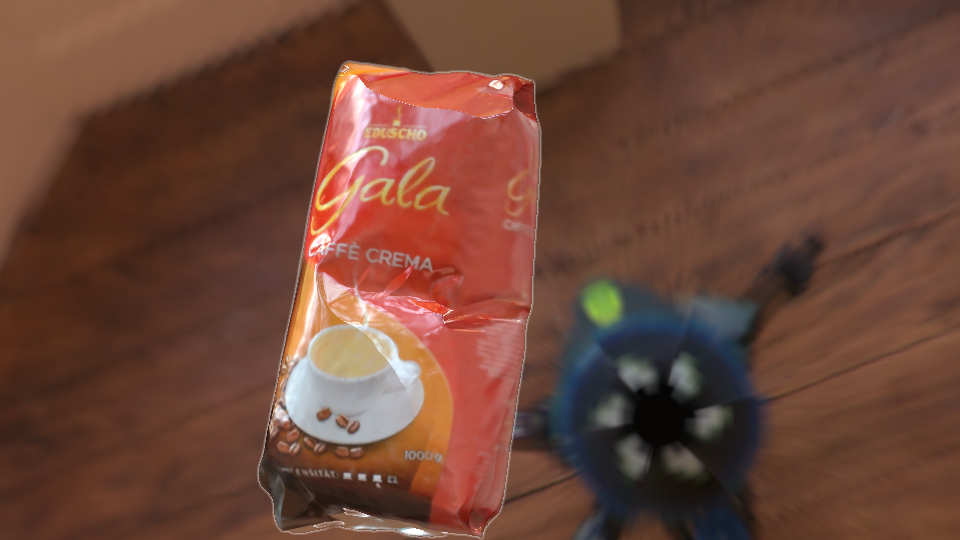}
        &
        \includegraphics[trim=150 0 290 65
        ,clip,width=0.1375\textwidth,angle=0]{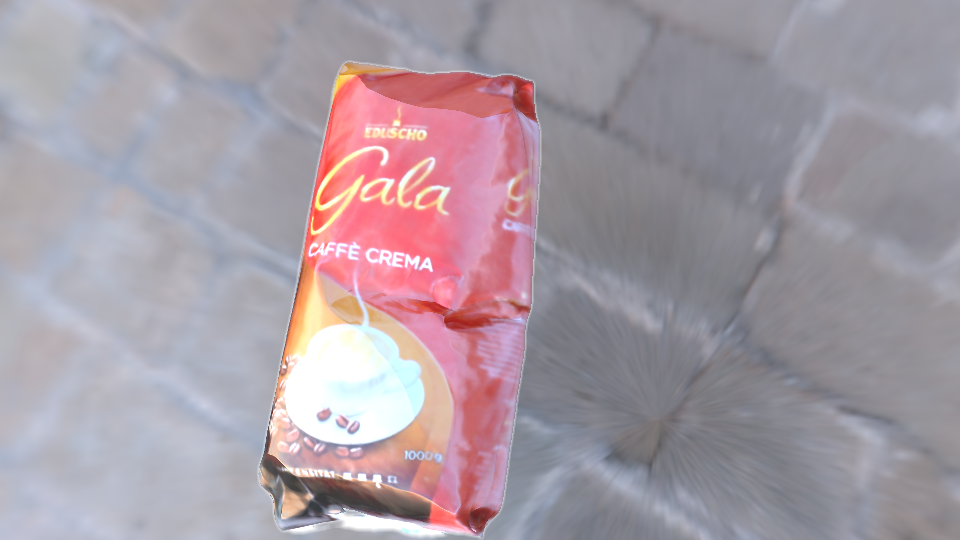}
        &
        \includegraphics[trim=150 0 290 65
        ,clip,width=0.1375\textwidth,angle=0]{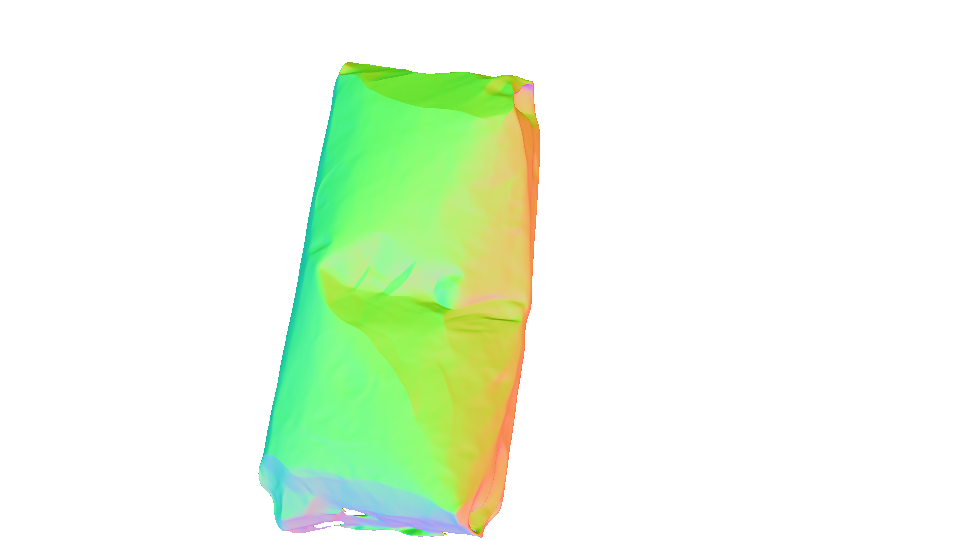}
        \\
        \vspace{-1.5em}
        \\
        \includegraphics[trim=215 145 165 140 ,clip,width=0.1375\textwidth,angle=0]{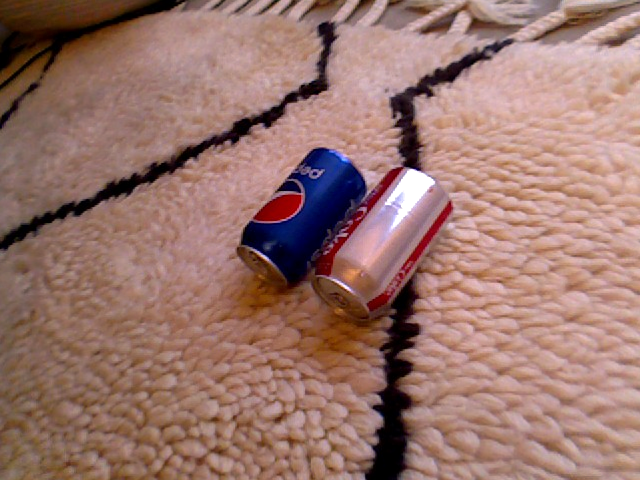}
        &
        \includegraphics[trim=215 145 165 140
        ,clip,width=0.1375\textwidth,angle=0]{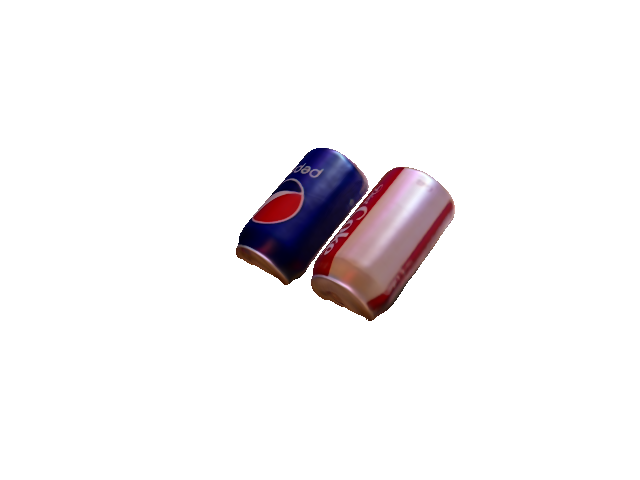}
        &
        \includegraphics[trim=215 145 165 140
        ,clip,width=0.1375\textwidth,angle=0]{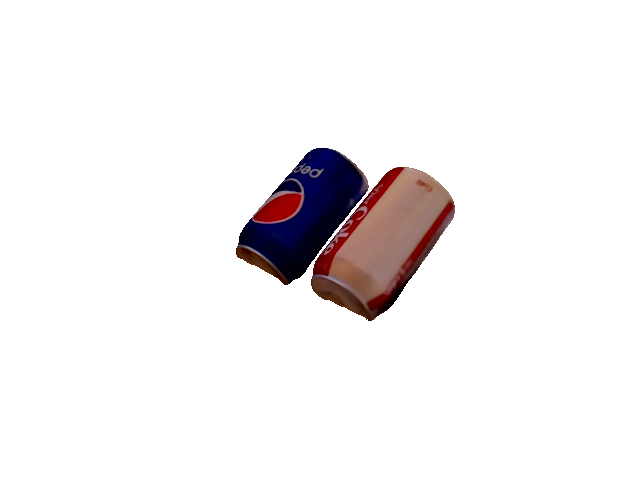}
        &
        \includegraphics[trim=215 145 165 140
        ,clip,width=0.1375\textwidth,angle=0]{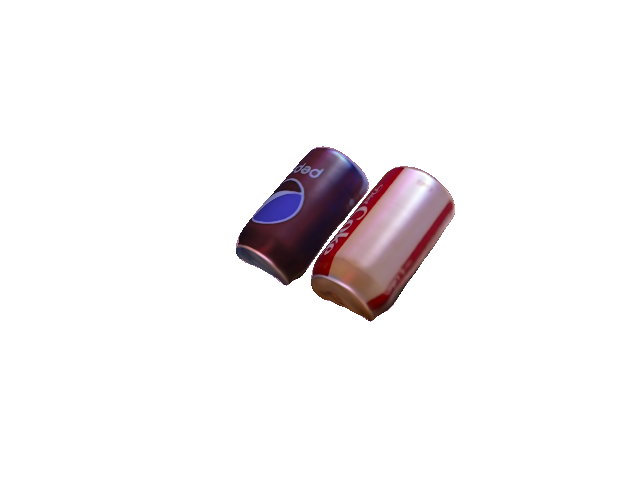}
        &   
        \includegraphics[trim=215 145 165 140,
        clip,width=0.1375\textwidth,angle=0]{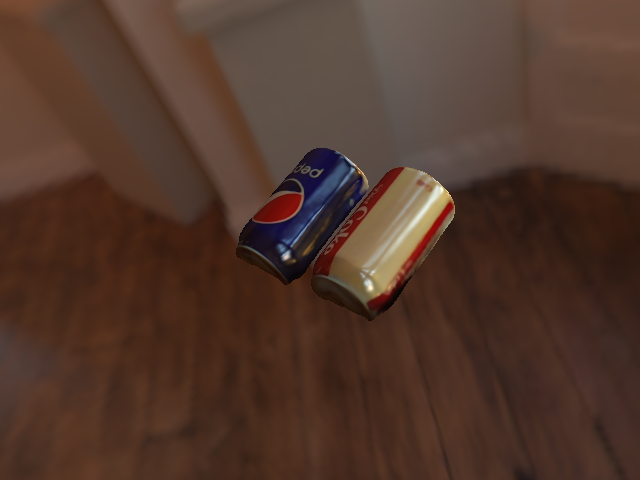}
        &
        \includegraphics[trim=215 145 165 140,
        clip,width=0.1375\textwidth,angle=0]{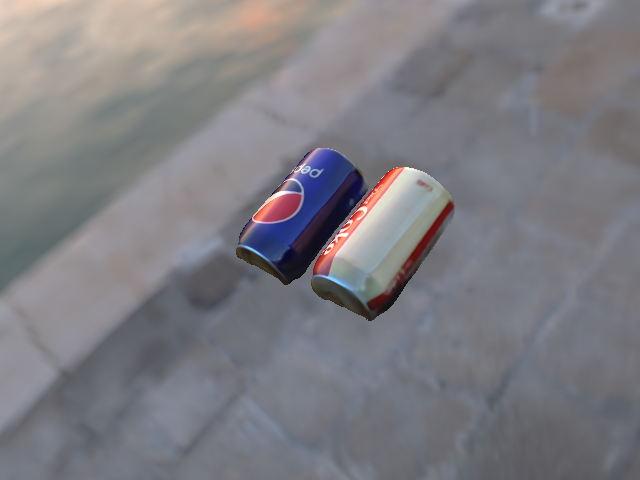}
        &
        \includegraphics[trim=215 145 165 140
        ,clip,width=0.1375\textwidth,angle=0]{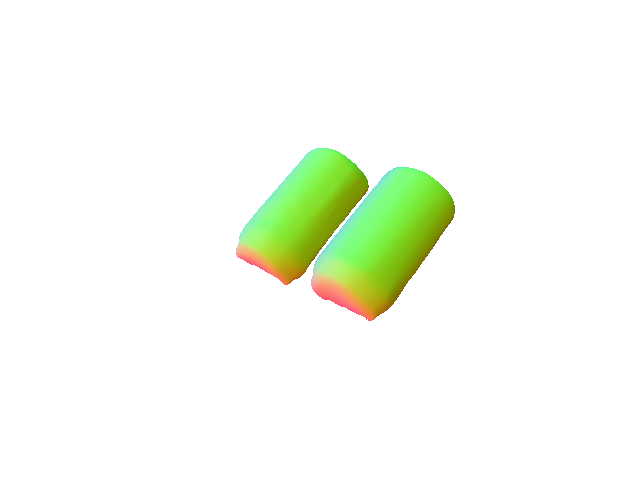}  
        \\
        \vspace{-1.5em}
        \\
        \includegraphics[trim=100 45 250 150 ,clip,width=0.1375\textwidth,angle=0]{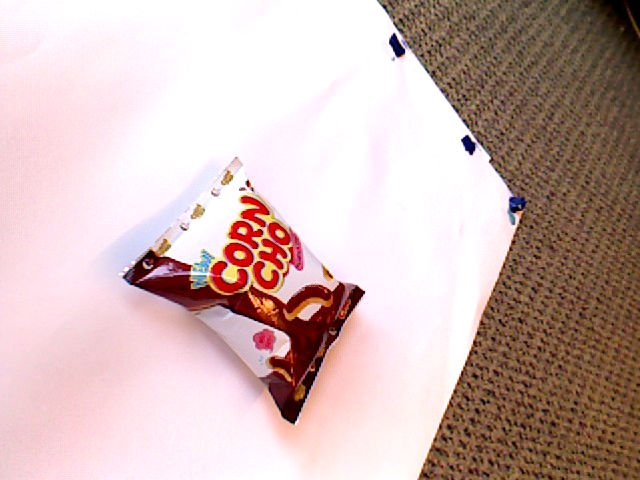}
        &
        \includegraphics[trim=100 45 250 150
        ,clip,width=0.1375\textwidth,angle=0]{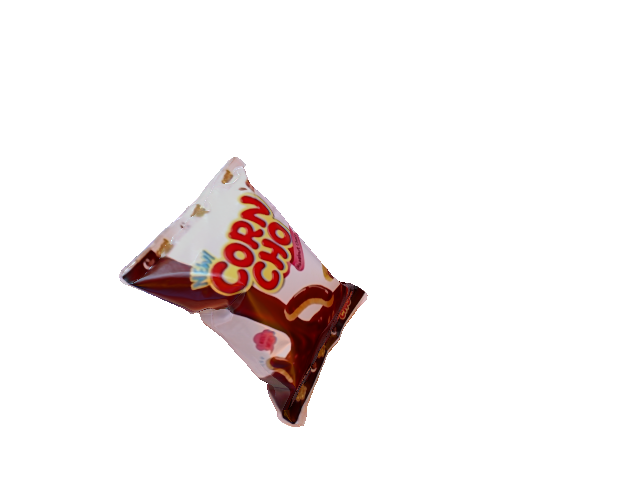}
        &
        \includegraphics[trim=100 45 250 150,clip,width=0.1375\textwidth,angle=0]{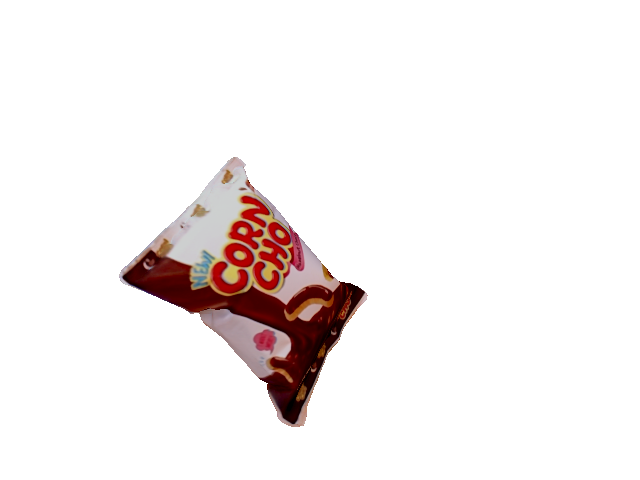}
        &
         \includegraphics[trim=100 45 250 150
        ,clip,width=0.1375\textwidth,angle=0]{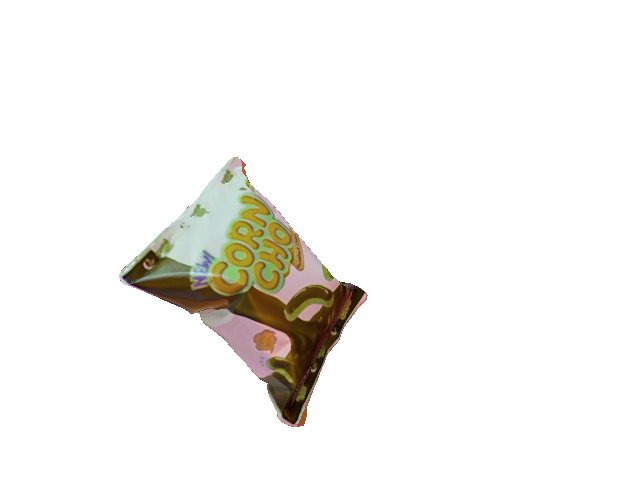}
        &   
        \includegraphics[trim=100 45 250 150,clip,width=0.1375\textwidth,angle=0]{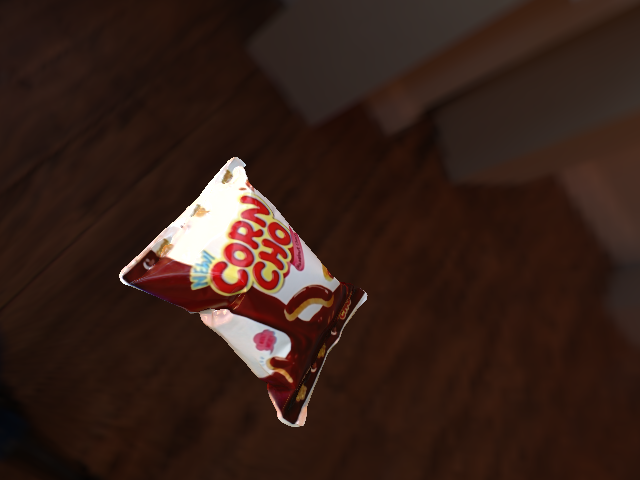}
        &
        \includegraphics[trim=100 45 250 150,clip,width=0.1375\textwidth,angle=0]{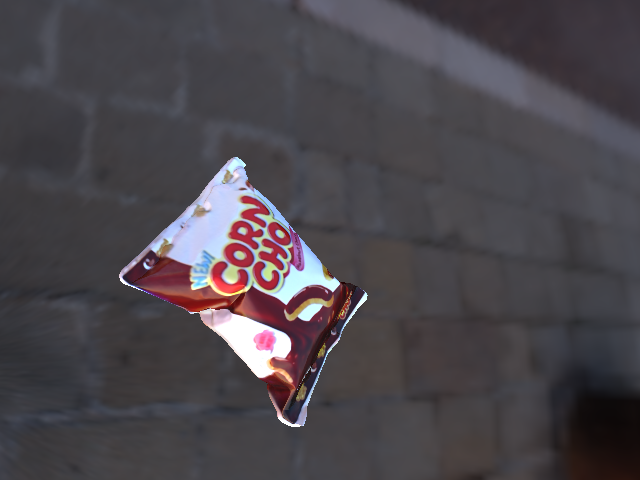}
        &
        \includegraphics[trim=100 45 250 150
        ,clip,width=0.1375\textwidth,angle=0]{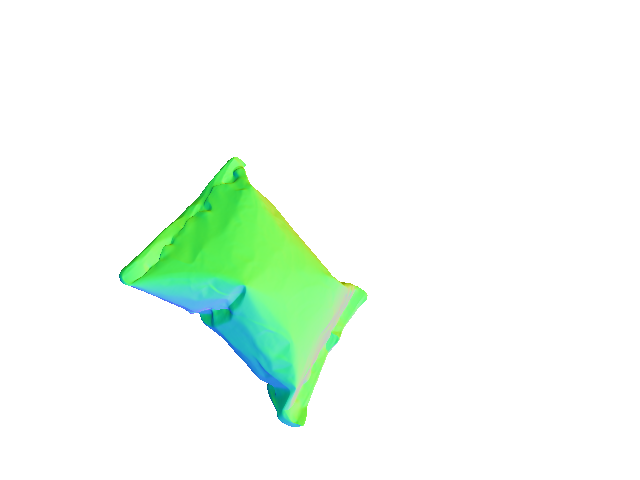}  
        \\
        \vspace{-1.5em}
        \\
        \includegraphics[trim=0 0 50 0
        ,clip,width=0.1375\textwidth,angle=0]{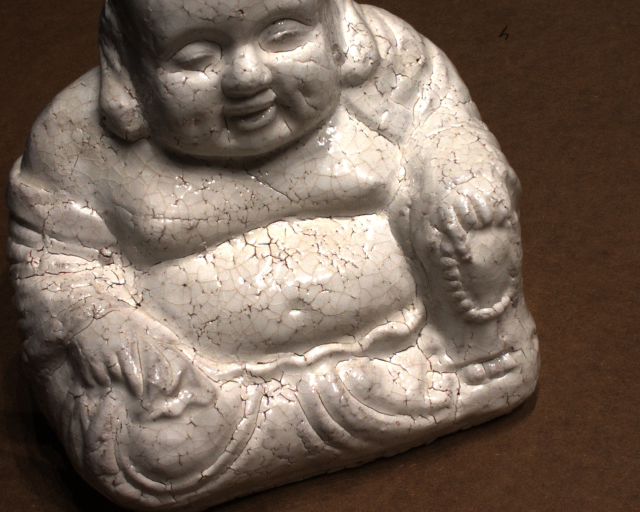}
        &
        \includegraphics[trim=0 0 50 0
        ,clip,width=0.1375\textwidth,angle=0]{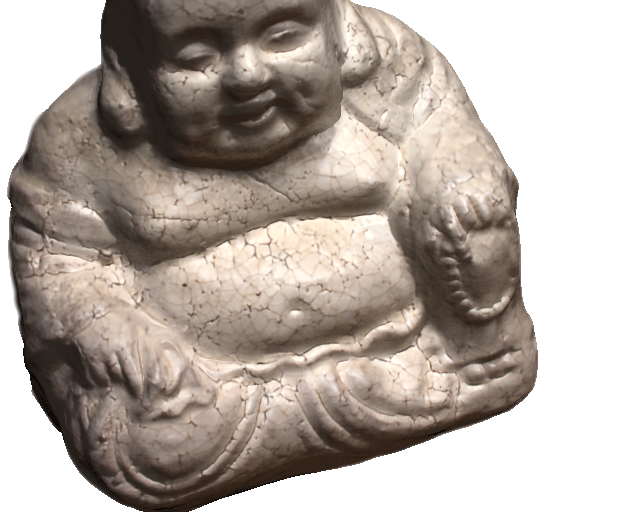}
        &
        \includegraphics[trim=0 0 50 0
        ,clip,width=0.1375\textwidth,angle=0]{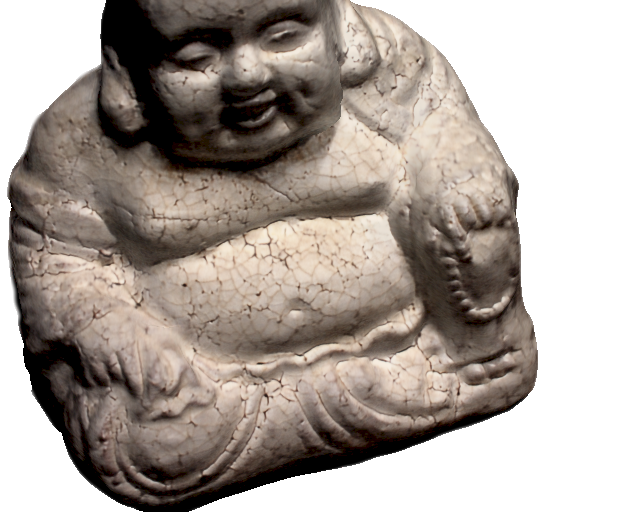}
        &
        \includegraphics[trim=0 0 50 0
        ,clip,width=0.1375\textwidth,angle=0]{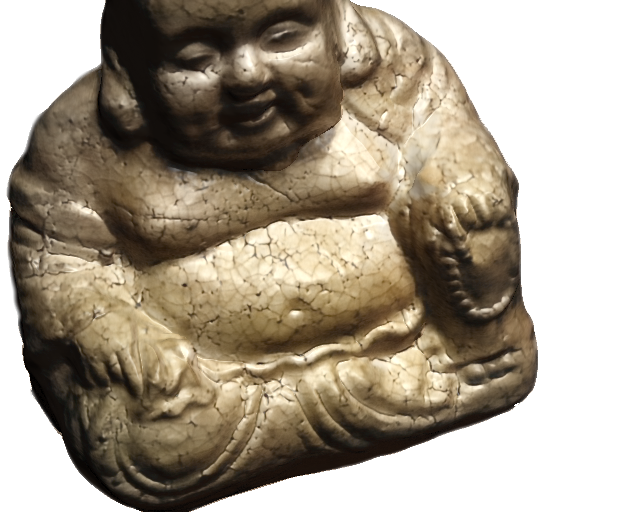}
        &   
        \includegraphics[trim=0 0 50 0
        ,clip,width=0.1375\textwidth,angle=0]{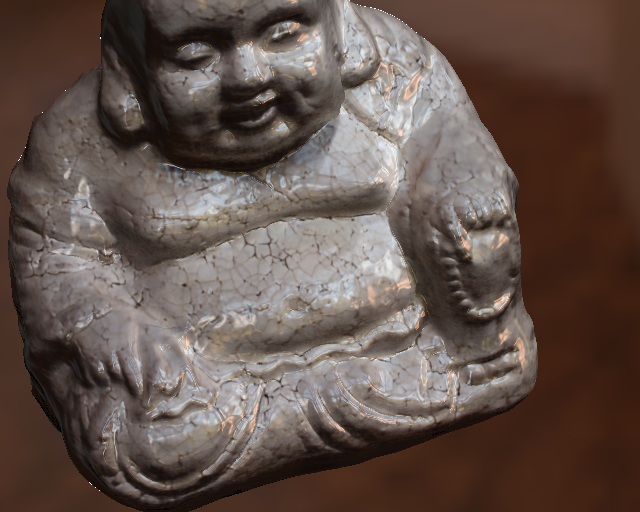}
        &
        \includegraphics[trim=0 0 50 0
        ,clip,width=0.1375\textwidth,angle=0]{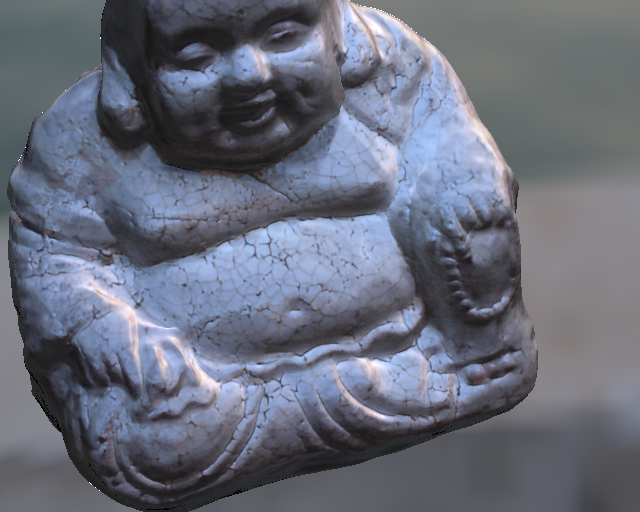}
        &
        \includegraphics[trim=0 0 50 0
        ,clip,width=0.1375\textwidth,angle=0]{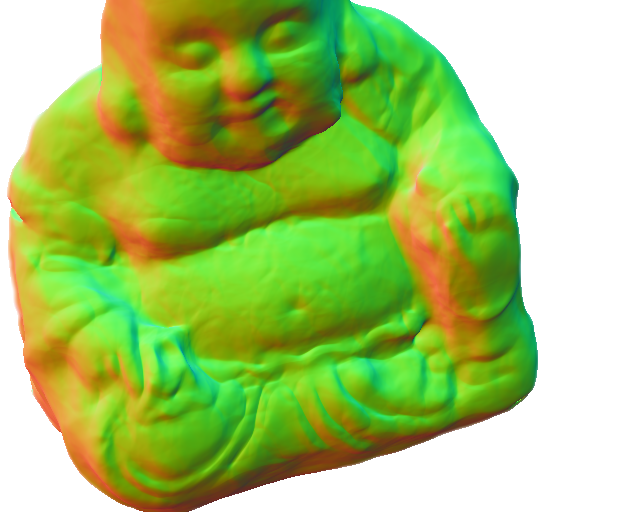}
     \end{tabular} 
  \vspace{-0.8\baselineskip}    
  \caption{With our pipeline, we can edit the materials and lighting of the real-world captures. For several input captures, we show from left to right: a real photo in the test set, our synthesized image, estimated diffuse image, editing results by painting diffuse albedo, relighting results under two novel environmental illuminations, and estimated surface normal. 
  }\label{fig:main_real}
\end{figure*}

%% file: sections/fig_envmap.tex
\begin{figure*}[htp]
\centering
  \includegraphics[trim=0 0 0 0,clip,width=0.96\textwidth]{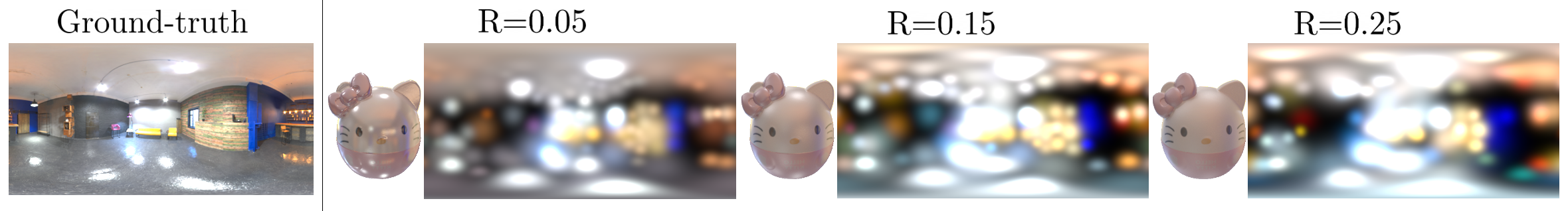}
  \vspace{-0.8\baselineskip}
  \caption{Ground truth and our reconstructed environment maps for the synthetic Kitty data with varying Ward BRDF roughness R. 
  For each roughness setting, an example training image rendered by Mitsuba~\cite{Mitsuba} is also shown. For rough surfaces (R=0.25), PhySG still recovers an environment map that resembles the ground truth, though blurrier. Nonetheless, this is sufficient to reconstruct the material accurately.
  }\label{fig:ablate_roughness}
  \vspace{-1.0\baselineskip}
\end{figure*}

%% file: sections/fig_main_syn.tex
\begin{figure*}[htp]
    \centering
    \setlength{\fboxsep}{0pt}%
    \setlength{\fboxrule}{0.5pt}%
    \contourlength{0.05em}%
    \vspace*{-2.2\baselineskip}%
    \hspace*{-2.0ex}%
    \begin{tikzpicture}[x=0.136\textwidth, y=0.115\textwidth,every text node part/.style={align=center}]
    \node[anchor=north west] at (0-0.02,  0) {\includegraphics[trim=100 80 100 120,clip, width=0.136\textwidth]{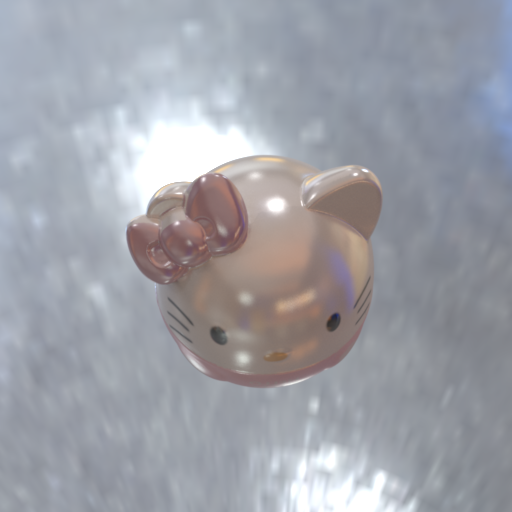}};
    \node[anchor=north west] at (1,       0) {\includegraphics[trim=100 80 100 120,clip,width=0.136\textwidth]{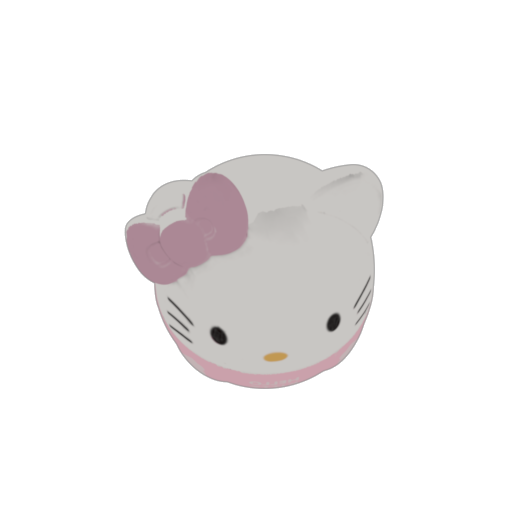}};
    \node[anchor=north west] at (2,       0) {\includegraphics[trim=100 80 100 120,clip,width=0.136\textwidth]{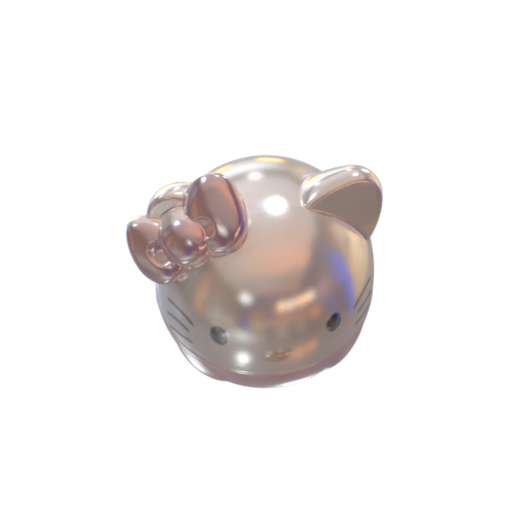}};
    \node[anchor=north west] at (3,       0) {\includegraphics[trim=100 80 100 120,clip,width=0.136\textwidth]{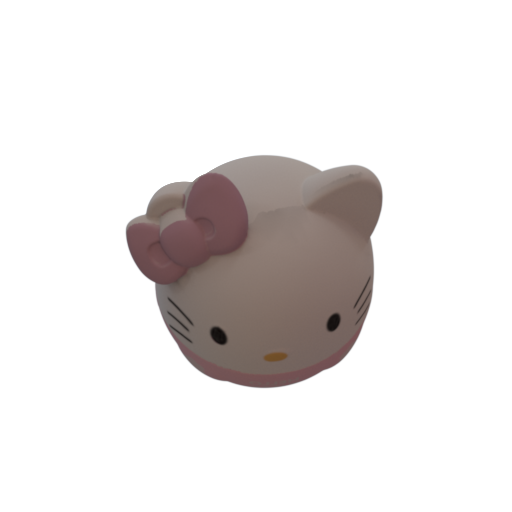}};
    \node[anchor=north west] at (4,       0) {\includegraphics[trim=100 80 100 120,clip,width=0.136\textwidth]{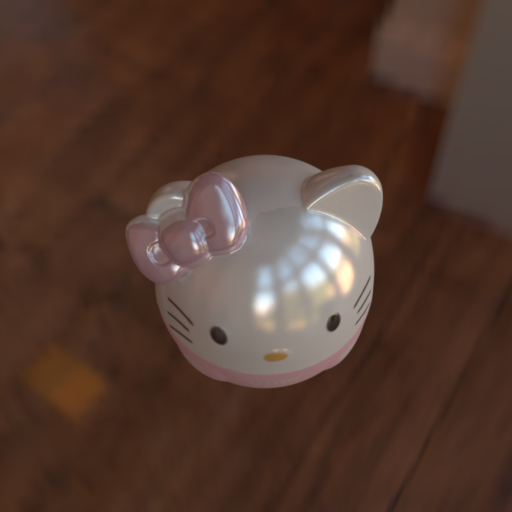}};
    \node[anchor=north west] at (5,       0) {\includegraphics[trim=100 80 100 120,clip,width=0.136\textwidth]{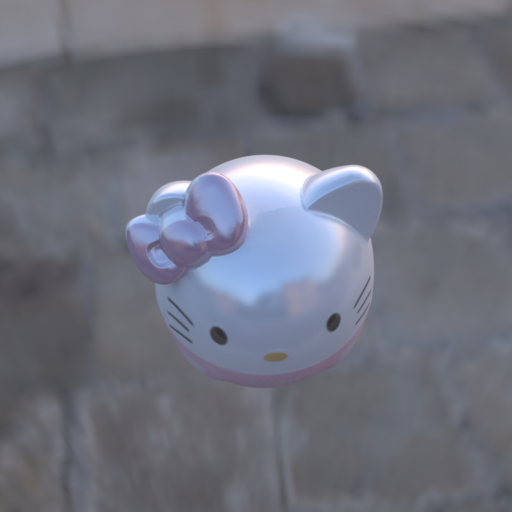}};
    \node[anchor=north west] at (6,       0) {\includegraphics[trim=100 80 100 120,clip,width=0.136\textwidth]{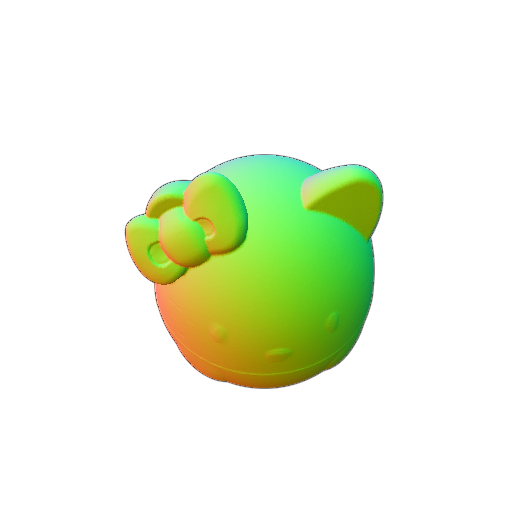}};
    \node[anchor=north west] at (0-0.02,  0- 1.11) {\includegraphics[trim=100 80 100 120,clip,width=0.136\textwidth]{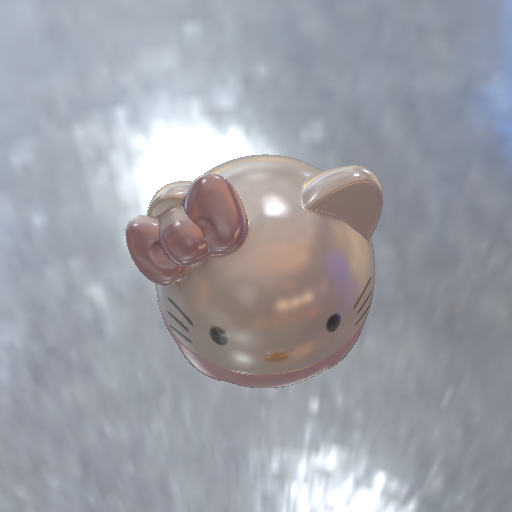}};
    \node[anchor=north west] at (1,       0- 1.11) {\includegraphics[trim=100 80 100 120,clip,width=0.136\textwidth]{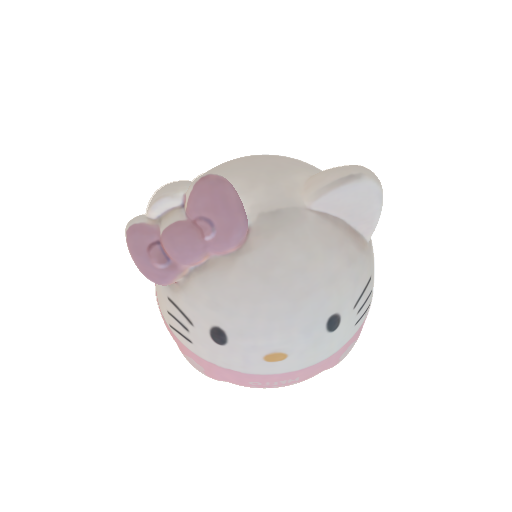}};
    \node[anchor=north west] at (2,       0- 1.11) {\includegraphics[trim=100 80 100 120,clip,width=0.136\textwidth]{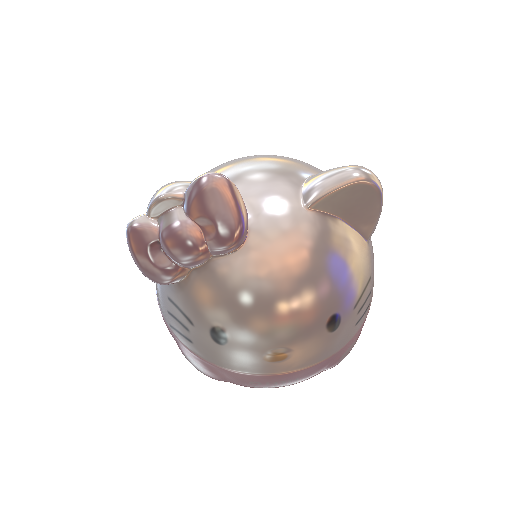}};
    \node[anchor=north west] at (3,       0- 1.11) {\includegraphics[trim=100 80 100 120,clip,width=0.136\textwidth]{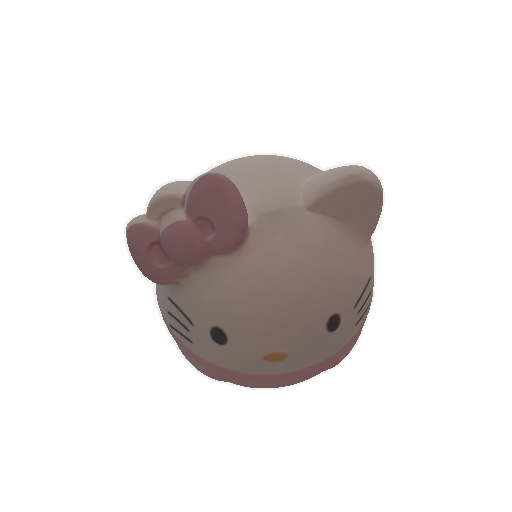}};
    \node[anchor=north west] at (4,       0- 1.11) {\includegraphics[trim=100 80 100 120,clip,width=0.136\textwidth]{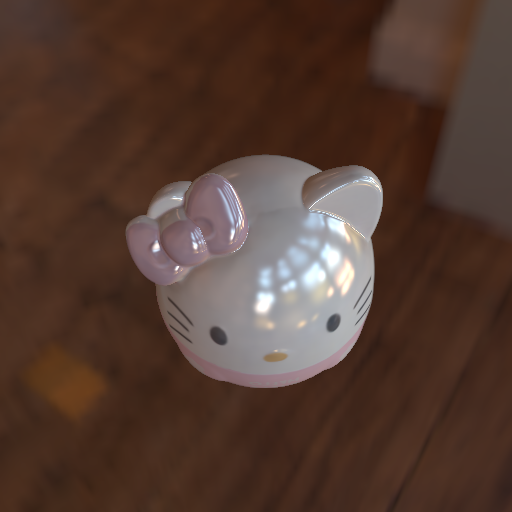}};
    \node[anchor=north west] at (5,       0- 1.11) {\includegraphics[trim=100 80 100 120,clip,width=0.136\textwidth]{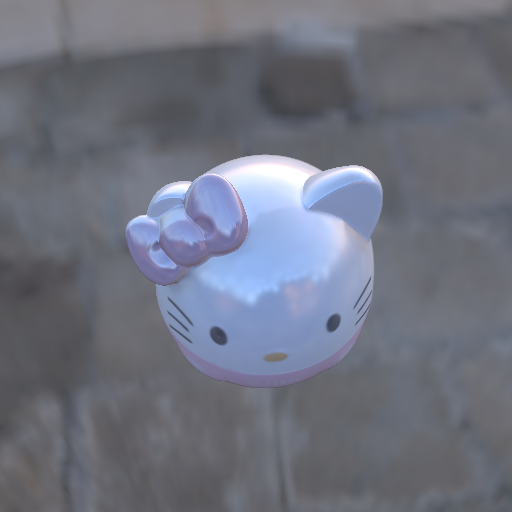}};
    \node[anchor=north west] at (6,       0- 1.11) {\includegraphics[trim=100 80 100 120,clip,width=0.136\textwidth]{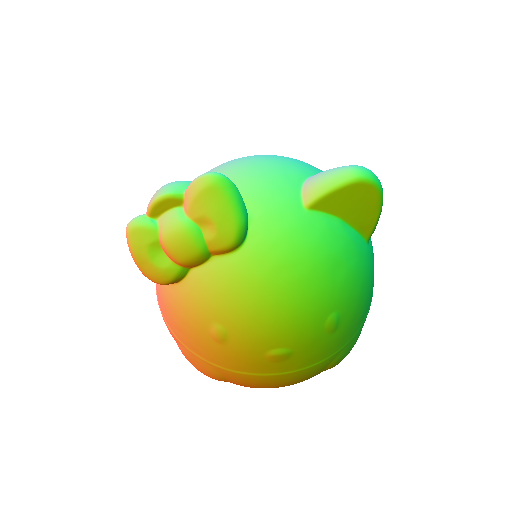}};
    \node[anchor=north west] at (0-0.02,  0-1.11*2) {\includegraphics[trim=100 80 100 120,clip, width=0.136\textwidth]{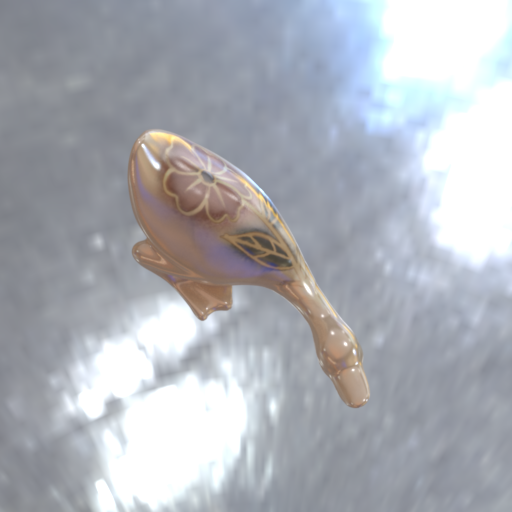}};
    \node[anchor=north west] at (1,       0-1.11*2) {\includegraphics[trim=100 80 100 120,clip,width=0.136\textwidth]{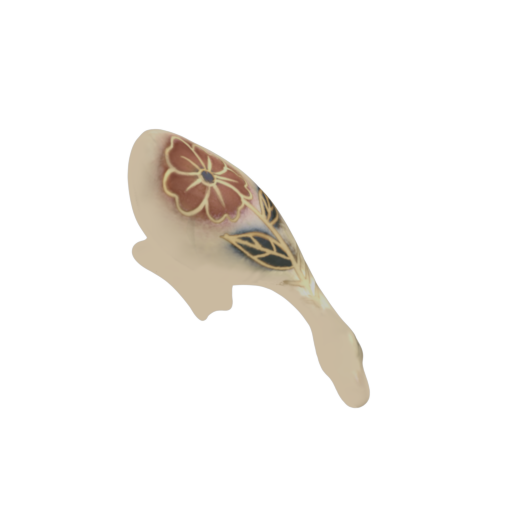}};
    \node[anchor=north west] at (2,       0-1.11*2) {\includegraphics[trim=100 80 100 120,clip,width=0.136\textwidth]{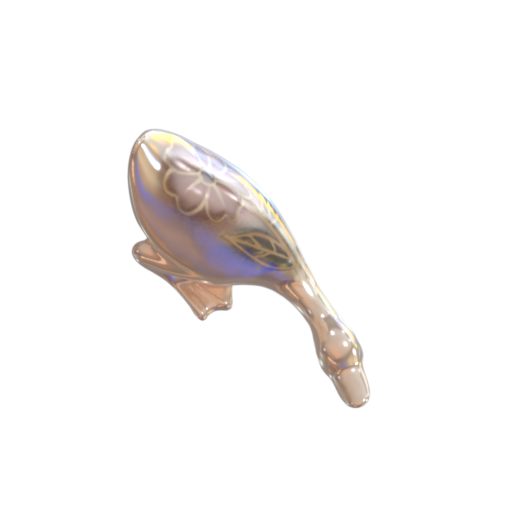}};
    \node[anchor=north west] at (3,       0-1.11*2) {\includegraphics[trim=100 80 100 120,clip,width=0.136\textwidth]{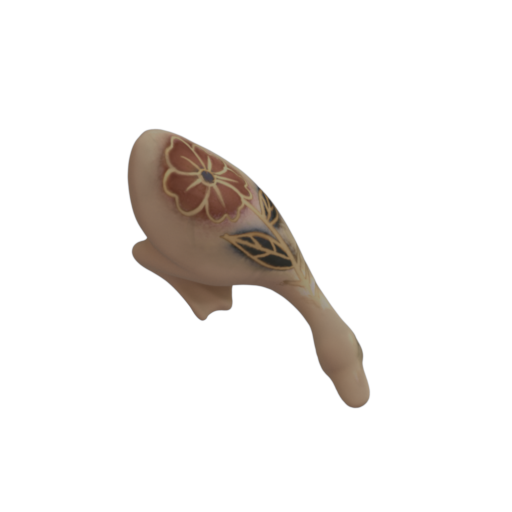}};
    \node[anchor=north west] at (4,       0-1.11*2) {\includegraphics[trim=100 80 100 120,clip,width=0.136\textwidth]{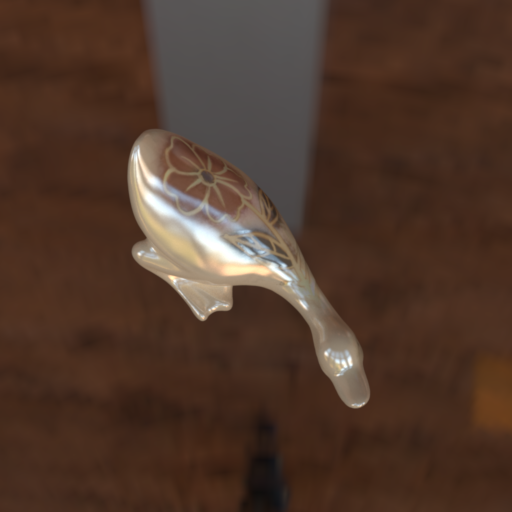}};
    \node[anchor=north west] at (5,       0-1.11*2) {\includegraphics[trim=100 80 100 120,clip,width=0.136\textwidth]{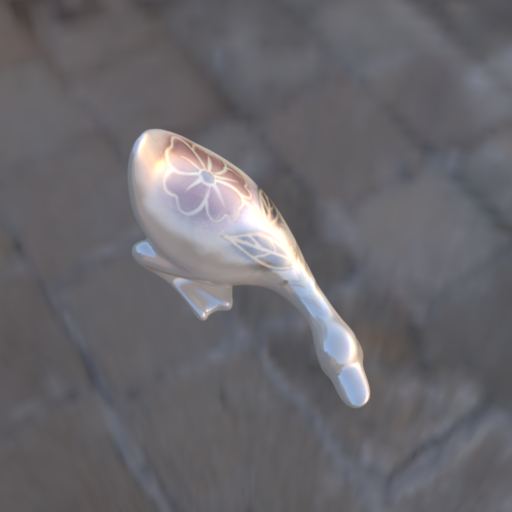}};
    \node[anchor=north west] at (6,       0-1.11*2) {\includegraphics[trim=100 80 100 120,clip,width=0.136\textwidth]{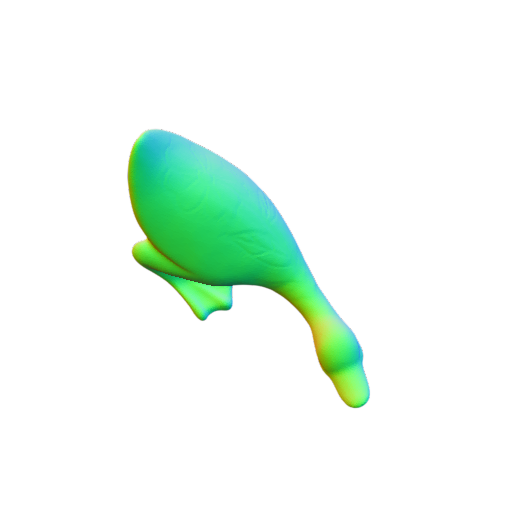}};
    \node[anchor=north west] at (0-0.02,  0- 1.11*3) {\includegraphics[trim=100 80 100 120,clip,width=0.136\textwidth]{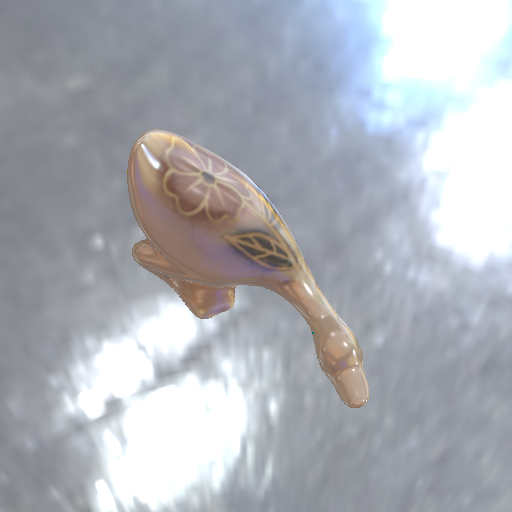}};
    \node[anchor=north west] at (1,       0- 1.11*3) {\includegraphics[trim=100 80 100 120,clip,width=0.136\textwidth]{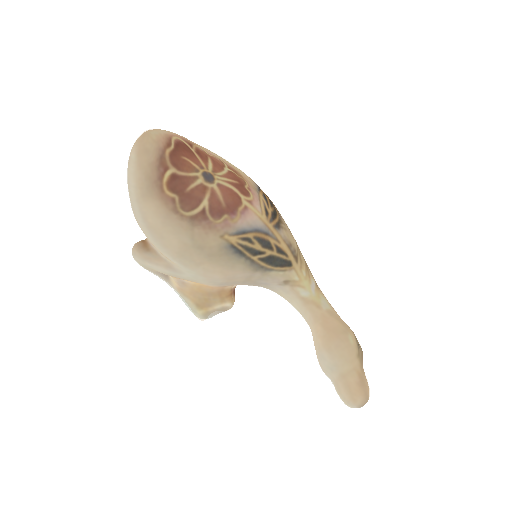}};
    \node[anchor=north west] at (2,       0- 1.11*3) {\includegraphics[trim=100 80 100 120,clip,width=0.136\textwidth]{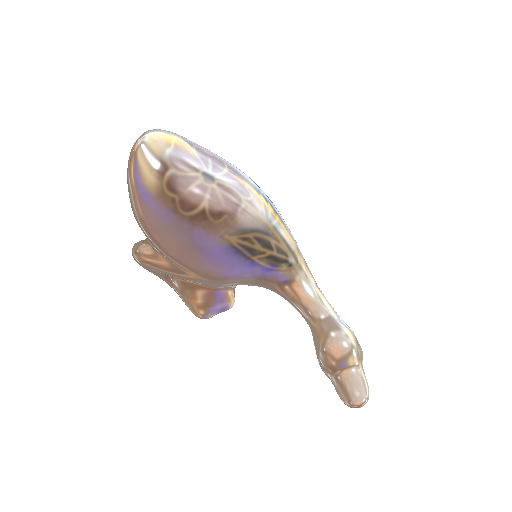}};
    \node[anchor=north west] at (3,       0- 1.11*3) {\includegraphics[trim=100 80 100 120,clip,width=0.136\textwidth]{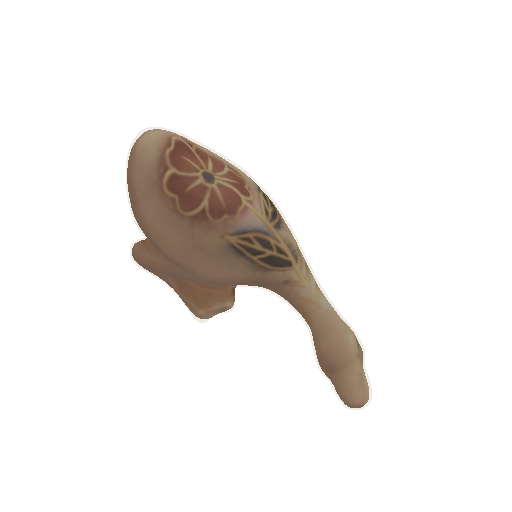}};
    \node[anchor=north west] at (4,       0- 1.11*3) {\includegraphics[trim=100 80 100 120,clip,width=0.136\textwidth]{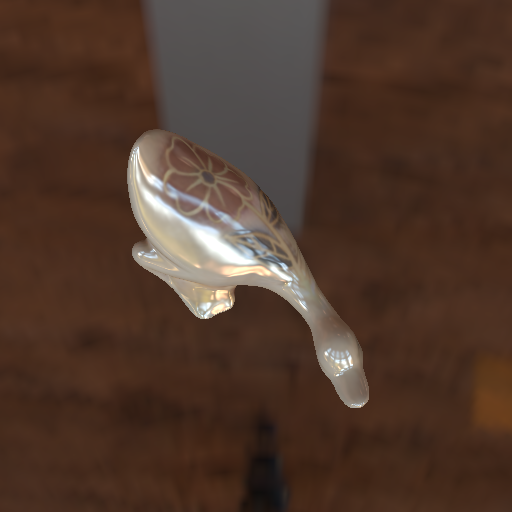}};
    \node[anchor=north west] at (5,       0- 1.11*3) {\includegraphics[trim=100 80 100 120,clip,width=0.136\textwidth]{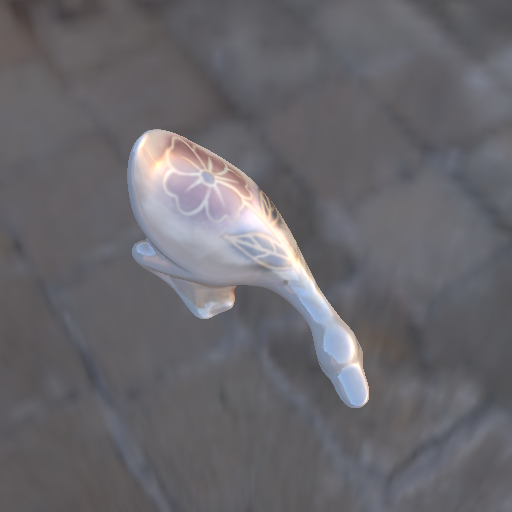}};
    \node[anchor=north west] at (6,       0- 1.11*3) {\includegraphics[trim=100 80 100 120,clip,width=0.136\textwidth]{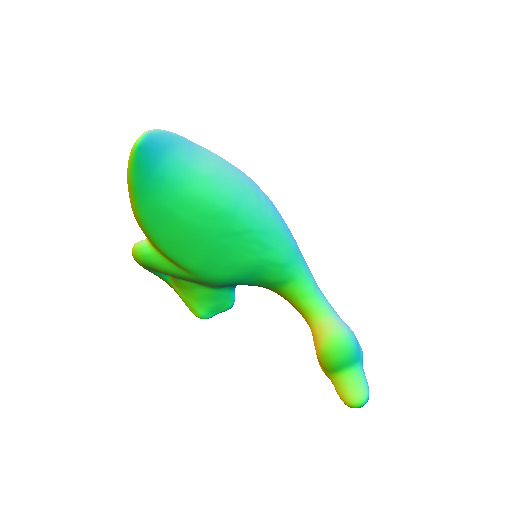}};
    \node[anchor=north west] at (0-0.02,  0-1.11*4) {\includegraphics[trim=100 100 100 100,clip, width=0.136\textwidth]{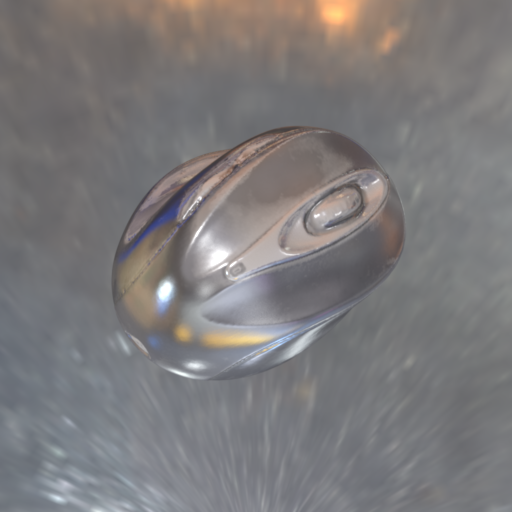}};
    \node[anchor=north west] at (1,       0-1.11*4) {\includegraphics[trim=100 100 100 100,clip,width=0.136\textwidth]{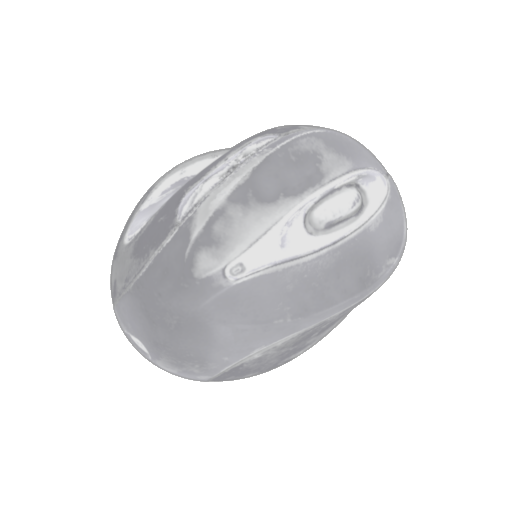}};
    \node[anchor=north west] at (2,       0-1.11*4) {\includegraphics[trim=100 100 100 100,clip,width=0.136\textwidth]{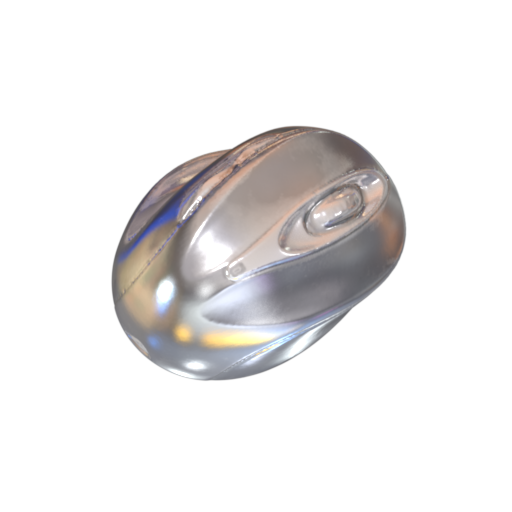}};
    \node[anchor=north west] at (3,       0-1.11*4) {\includegraphics[trim=100 100 100 100,clip,width=0.136\textwidth]{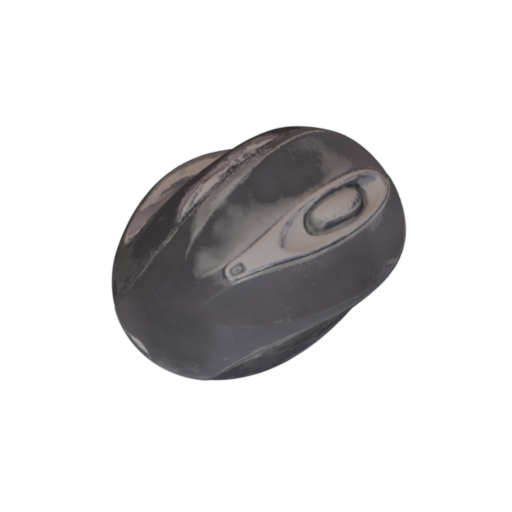}};
    \node[anchor=north west] at (4,       0-1.11*4) {\includegraphics[trim=100 100 100 100,clip,width=0.136\textwidth]{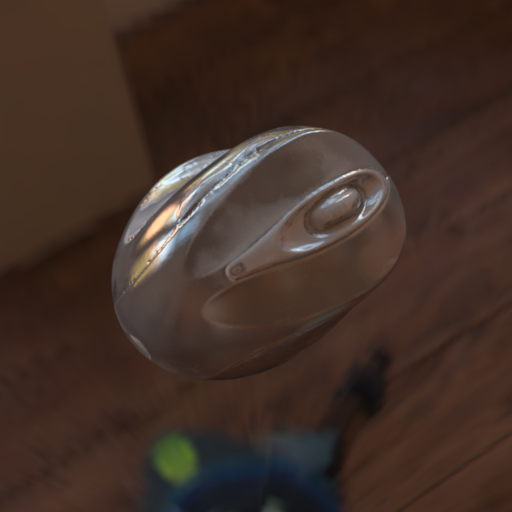}};
    \node[anchor=north west] at (5,       0-1.11*4) {\includegraphics[trim=100 100 100 100,clip,width=0.136\textwidth]{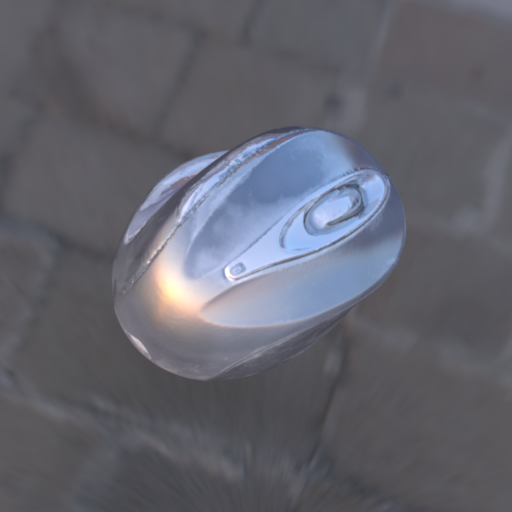}};
    \node[anchor=north west] at (6,       0-1.11*4) {\includegraphics[trim=100 100 100 100,clip,width=0.136\textwidth]{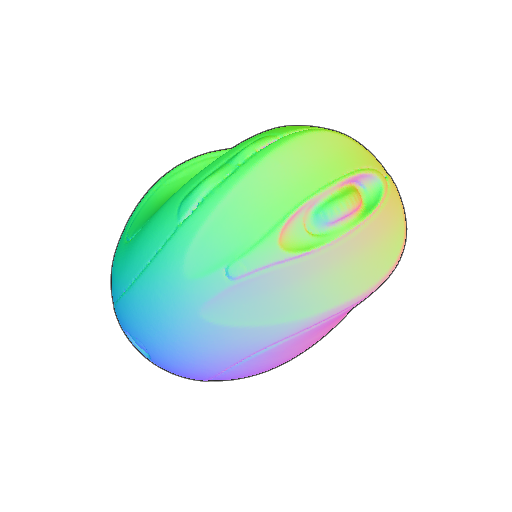}};
    \node[anchor=north west] at (0-0.02,  0- 1.11*5) {\includegraphics[trim=100 100 100 100,clip,width=0.136\textwidth]{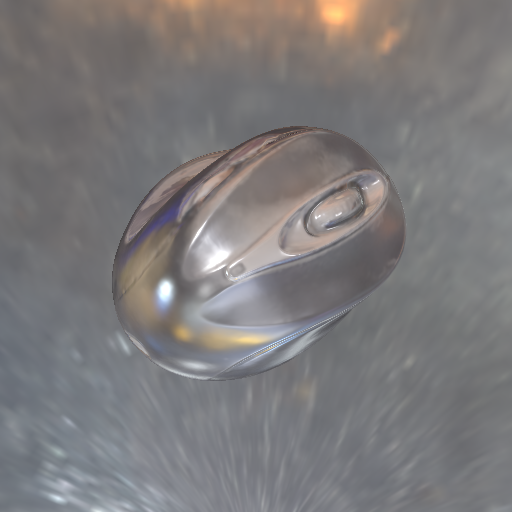}};
    \node[anchor=north west] at (1,       0- 1.11*5) {\includegraphics[trim=100 100 100 100,clip,width=0.136\textwidth]{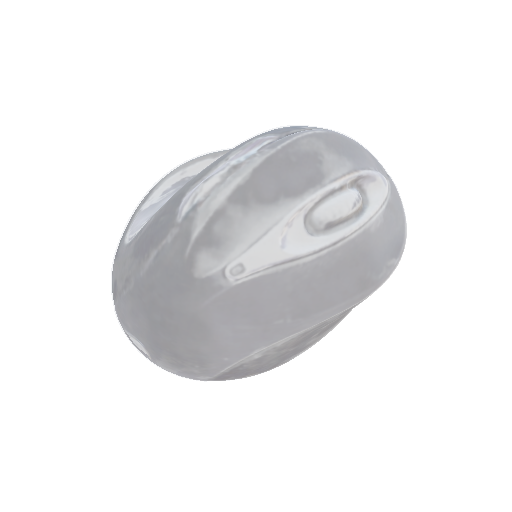}};
    \node[anchor=north west] at (2,       0- 1.11*5) {\includegraphics[trim=100 100 100 100,clip,width=0.136\textwidth]{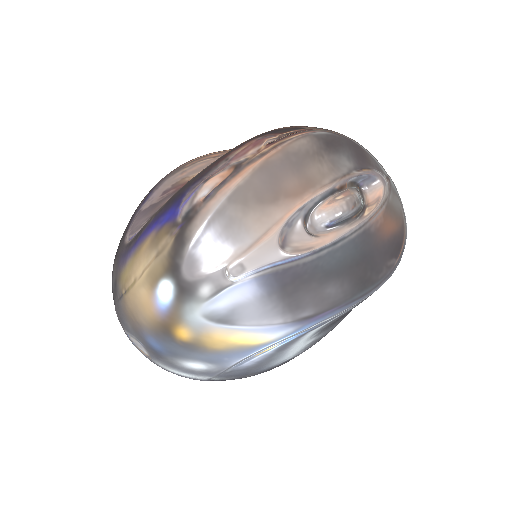}};
    \node[anchor=north west] at (3,       0- 1.11*5) {\includegraphics[trim=100 100 100 100,clip,width=0.136\textwidth]{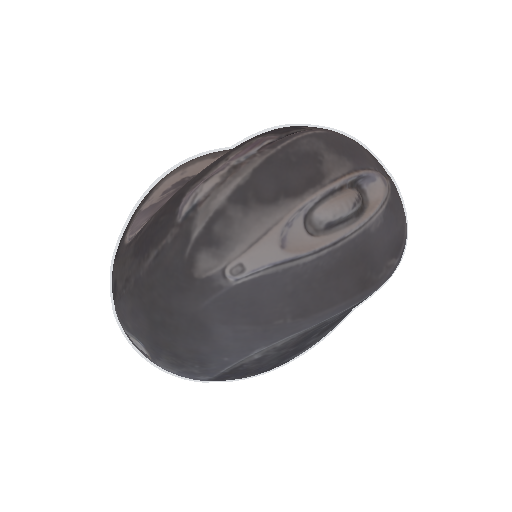}};
    \node[anchor=north west] at (4,       0- 1.11*5) {\includegraphics[trim=100 100 100 100,clip,width=0.136\textwidth]{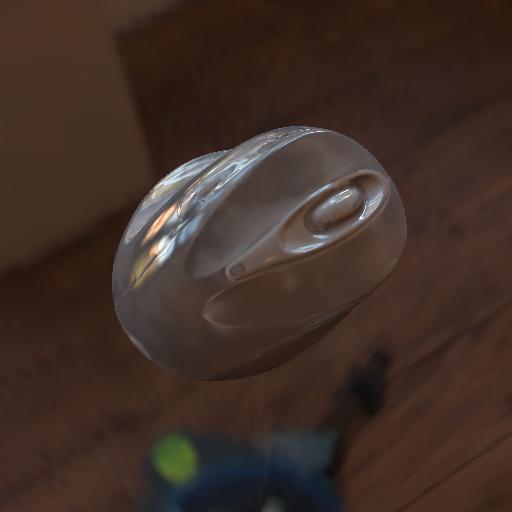}};
    \node[anchor=north west] at (5,       0- 1.11*5) {\includegraphics[trim=100 100 100 100,clip,width=0.136\textwidth]{
 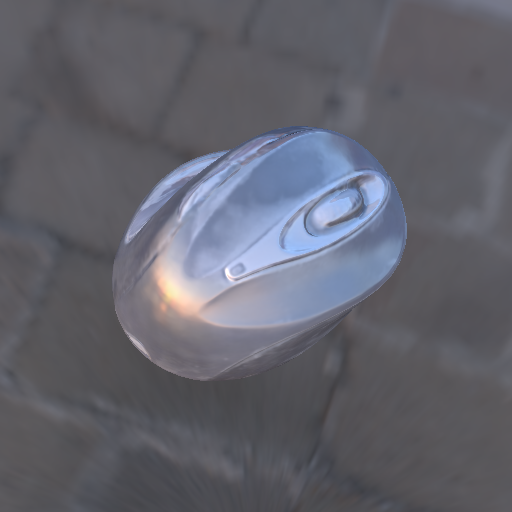}};
    \node[anchor=north west] at (6,       0- 1.11*5) {\includegraphics[trim=100 100 100 100,clip,width=0.136\textwidth]{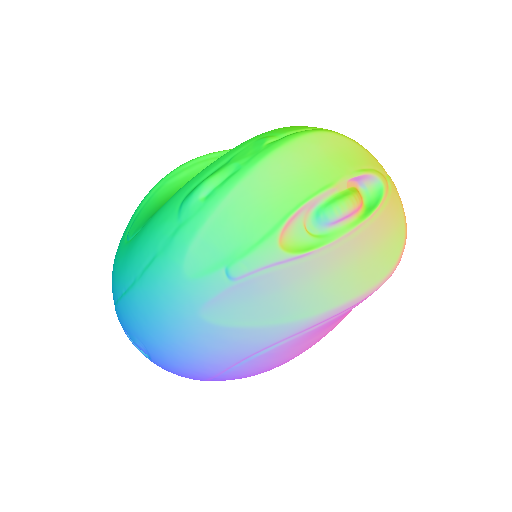}};
    \node [rotate=90] at (-0.05, -0.59)   {\textsc{\small GT}};
    \node [rotate=90] at (-0.05, -1.7)   {\textsc{\small Ours}};
    
    \node [rotate=90] at (-0.05,  -0.58 - 2.23)   {\textsc{\small GT}};
    \node [rotate=90] at (-0.05, -1.7 - 2.23)   {\textsc{\small Ours}};
    
    \node [rotate=90] at (-0.05,  -0.58 - 2.23*2)   {\textsc{\small GT}};
    \node [rotate=90] at (-0.05, -1.7 - 2.25*2)   {\textsc{\small Ours}};

    \node[anchor=north west] at (0.13,   0.15) {\small Novel-view};
    \node[anchor=north west] at (1.1,    0.15) {\small Diffuse albedo};
    \node[anchor=north west] at (2.38,   0.15) {\small Edit 1};
    \node[anchor=north west] at (3.38,   0.15) {\small Edit 2};
    \node[anchor=north west] at (4.25,   0.15) {\small Relight 1};
    \node[anchor=north west] at (5.25,   0.15) {\small Relight 2};
    \node[anchor=north west] at (6.1,    0.15) {\small Surface normal};
    \end{tikzpicture}%
    \caption{Results of our pipeline on synthetic data. For a novel test view, we compare our predicted image, estimated diffuse albedo, specular BRDF editing results and relighting results to ground truth images rendered by Mitsuba~\cite{Mitsuba}. Note that there is a scale ambiguity in inverse rendering problems; hence we align our estimated diffuse albedo to the ground truth for visualization here (see Eq.~\ref{eq:align_scale} for details of the alignment we apply). More examples are available in the supplemental material.
    }
    \label{fig:main_syn}
    \vspace{-1.0\baselineskip}
\end{figure*}
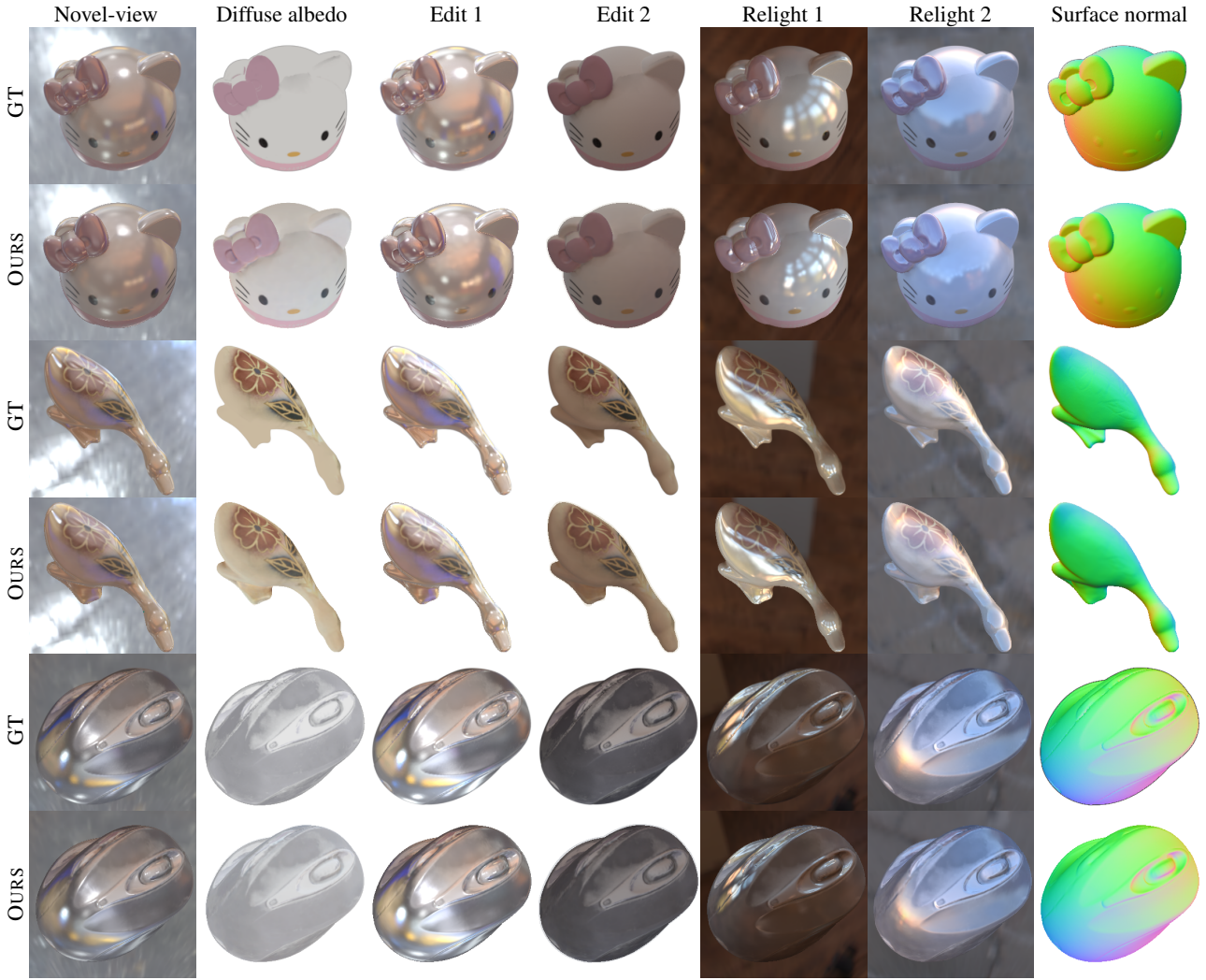

%% file: sections/05-conclusion.tex
\section{Conclusion}
We proposed PhySG, an end-the-end inverse rendering pipeline that uses physics-based differentiable rendering. PhySG uses signed distance functions (SDFs) and spherical Gaussians (SG) to represent geometry and appearance, respectively. We show that PhySG can jointly recover environment maps, material BRDFs and geometry from multi-view inputs captured under static illumination, enabling physics-based material editing and relighting.

\medskip
\noindent\textbf{Limitations.} Our method has a few limitations that can be the subject of future work.  First, indirect illumination is not modelled by our SG approximation of the rendering equation, which limits our method to object-level data. To lift this restriction and extend to scene-level data, differentiable path tracing combined with deferred neural textures~\cite{gao2020deferred,thies2019deferred} can be explored, where only rough geometry is required for guidance. Second, we assume constant and monochrome specular BRDFs (with spatially-varying diffuse components). This assumption is due to the scale ambiguity between illumination and reflectance. Similar to intrinsic image decomposition, learning-based priors could help alleviate such ambiguities.
Last, our work can also be extended to handle anisotropic or data-driven BRDFs, e.g., by fitting a mixture of anisotropic SGs~\cite{xu2013anisotropic}.

\medskip
\noindent \textbf{Acknowledgements.} This work was supported in part by the National Science Foundation (IIS-2008313, CHS-1900783, CHS-1930755).

%% file: main.bbl
\begin{thebibliography}{10}\itemsep=-1pt

\bibitem{remove_bg}
Remove image background.
\newblock \url{https://www.remove.bg/}.
\newblock Accessed: 2020-10-15.

\bibitem{aanaes2016large}
Henrik Aan{\ae}s, Rasmus~Ramsb{\o}l Jensen, George Vogiatzis, Engin Tola, and
  Anders~Bjorholm Dahl.
\newblock Large-scale data for multiple-view stereopsis.
\newblock {\em Int. J. Comput. Vis.}, pages 1--16, 2016.

\bibitem{ackermann2015survey}
Jens Ackermann and Michael Goesele.
\newblock A survey of photometric stereo techniques.
\newblock {\em Foundations and Trends{\textregistered} in Computer Graphics and
  Vision}, 9(3-4):149--254, 2015.

\bibitem{azinovic2019inverse}
Dejan Azinovic, Tzu-Mao Li, Anton Kaplanyan, and Matthias Nie{\ss}ner.
\newblock Inverse path tracing for joint material and lighting estimation.
\newblock In {\em IEEE Conf. Comput. Vis. Pattern Recog.}, pages 2447--2456,
  2019.

\bibitem{barron2014shape}
Jonathan~T Barron and Jitendra Malik.
\newblock Shape, illumination, and reflectance from shading.
\newblock {\em IEEE Trans. Pattern Anal. Mach. Intell.}, 37(8):1670--1687,
  2014.

\bibitem{bi2020neural}
Sai Bi, Zexiang Xu, Pratul Srinivasan, Ben Mildenhall, Kalyan Sunkavalli,
  Milo{\v{s}} Ha{\v{s}}an, Yannick Hold-Geoffroy, David Kriegman, and Ravi
  Ramamoorthi.
\newblock Neural reflectance fields for appearance acquisition.
\newblock {\em arXiv preprint arXiv:2008.03824}, 2020.

\bibitem{bi2020deepreflectance}
Sai Bi, Zexiang Xu, Kalyan Sunkavalli, Milo{\v{s}} Ha{\v{s}}an, Yannick
  Hold-Geoffroy, David Kriegman, and Ravi Ramamoorthi.
\newblock Deep reflectance volumes: Relightable reconstructions from multi-view
  photometric images.
\newblock {\em arXiv preprint arXiv:2007.09892}, 2020.

\bibitem{bi2020deep}
Sai Bi, Zexiang Xu, Kalyan Sunkavalli, David Kriegman, and Ravi Ramamoorthi.
\newblock Deep 3d capture: Geometry and reflectance from sparse multi-view
  images.
\newblock In {\em IEEE Conf. Comput. Vis. Pattern Recog.}, pages 5960--5969,
  2020.

\bibitem{burley2012physically}
Brent Burley and Walt Disney~Animation Studios.
\newblock Physically-based shading at disney.
\newblock In {\em SIGGRAPH}, volume 2012, pages 1--7. vol. 2012, 2012.

\bibitem{dong2014appearance}
Yue Dong, Guojun Chen, Pieter Peers, Jiawan Zhang, and Xin Tong.
\newblock Appearance-from-motion: Recovering spatially varying surface
  reflectance under unknown lighting.
\newblock {\em ACM Trans. Graph.}, 33(6):1--12, 2014.

\bibitem{gao2020deferred}
Duan Gao, Guojun Chen, Yue Dong, Pieter Peers, Kun Xu, and Xin Tong.
\newblock Deferred neural lighting: free-viewpoint relighting from unstructured
  photographs.
\newblock {\em ACM Transactions on Graphics (TOG)}, 39(6):1--15, 2020.

\bibitem{gropp2020implicit}
Amos Gropp, Lior Yariv, Niv Haim, Matan Atzmon, and Yaron Lipman.
\newblock Implicit geometric regularization for learning shapes.
\newblock {\em arXiv preprint arXiv:2002.10099}, 2020.

\bibitem{haber2009relighting}
Tom Haber, Christian Fuchs, Philippe Bekaer, Hans-Peter Seidel, Michael
  Goesele, and Hendrik~PA Lensch.
\newblock Relighting objects from image collections.
\newblock In {\em IEEE Conf. Comput. Vis. Pattern Recog.}, pages 627--634.
  IEEE, 2009.

\bibitem{Mitsuba}
Wenzel Jakob.
\newblock Mitsuba renderer, 2010.
\newblock http://www.mitsuba-renderer.org.

\bibitem{jiang2020sdfdiff}
Yue Jiang, Dantong Ji, Zhizhong Han, and Matthias Zwicker.
\newblock Sdfdiff: Differentiable rendering of signed distance fields for 3d
  shape optimization.
\newblock In {\em IEEE Conf. Comput. Vis. Pattern Recog.}, pages 1251--1261,
  2020.

\bibitem{kajiya1986rendering}
James~T Kajiya.
\newblock The rendering equation.
\newblock In {\em Proceedings of the 13th annual conference on Computer
  graphics and interactive techniques}, pages 143--150, 1986.

\bibitem{karis2013real}
Brian Karis and Epic Games.
\newblock Real shading in unreal engine 4.

\bibitem{keinert2015spherical}
Benjamin Keinert, Matthias Innmann, Michael S{\"a}nger, and Marc Stamminger.
\newblock Spherical fibonacci mapping.
\newblock {\em ACM Trans. Graph.}, 34(6):1--7, 2015.

\bibitem{Li:2018:DMC}
Tzu-Mao Li, Miika Aittala, Fr{\'e}do Durand, and Jaakko Lehtinen.
\newblock Differentiable monte carlo ray tracing through edge sampling.
\newblock {\em ACM Trans. Graph. (Proc. SIGGRAPH Asia)}, 37(6):222:1--222:11,
  2018.

\bibitem{li2020inverse}
Zhengqin Li, Mohammad Shafiei, Ravi Ramamoorthi, Kalyan Sunkavalli, and
  Manmohan Chandraker.
\newblock Inverse rendering for complex indoor scenes: Shape, spatially-varying
  lighting and svbrdf from a single image.
\newblock In {\em Proceedings of the IEEE/CVF Conference on Computer Vision and
  Pattern Recognition}, pages 2475--2484, 2020.

\bibitem{li2020crowdsampling}
Zhengqi Li, Wenqi Xian, Abe Davis, and Noah Snavely.
\newblock Crowdsampling the plenoptic function.
\newblock In {\em Eur. Conf. Comput. Vis.}, 2020.

\bibitem{li2018learning}
Zhengqin Li, Zexiang Xu, Ravi Ramamoorthi, Kalyan Sunkavalli, and Manmohan
  Chandraker.
\newblock Learning to reconstruct shape and spatially-varying reflectance from
  a single image.
\newblock {\em ACM Trans. Graph.}, 37(6):1--11, 2018.

\bibitem{lombardi2016radiometric}
Stephen Lombardi and Ko Nishino.
\newblock Radiometric scene decomposition: Scene reflectance, illumination, and
  geometry from rgb-d images.
\newblock In {\em 3DV}, pages 305--313. IEEE, 2016.

\bibitem{maier2017intrinsic3d}
Robert Maier, Kihwan Kim, Daniel Cremers, Jan Kautz, and Matthias Nie{\ss}ner.
\newblock {Intrinsic3D}: High-quality {3D} reconstruction by joint appearance
  and geometry optimization with spatially-varying lighting.
\newblock In {\em Int. Conf. Comput. Vis.}, 2017.

\bibitem{martin2020nerf}
Ricardo Martin-Brualla, Noha Radwan, Mehdi~SM Sajjadi, Jonathan~T Barron,
  Alexey Dosovitskiy, and Daniel Duckworth.
\newblock Nerf in the wild: Neural radiance fields for unconstrained photo
  collections.
\newblock {\em arXiv preprint arXiv:2008.02268}, 2020.

\bibitem{hemisphere_int}
Julian Meder and Beat Br{\"u}derlin.
\newblock Hemispherical gaussians for accurate light integration.
\newblock In Leszek~J. Chmielewski, Ryszard Kozera, Arkadiusz Or{\l}owski,
  Konrad Wojciechowski, Alfred~M. Bruckstein, and Nicolai Petkov, editors, {\em
  Computer Vision and Graphics}, pages 3--15, Cham, 2018. Springer
  International Publishing.

\bibitem{mescheder2019occupancy}
Lars Mescheder, Michael Oechsle, Michael Niemeyer, Sebastian Nowozin, and
  Andreas Geiger.
\newblock Occupancy networks: Learning 3d reconstruction in function space.
\newblock In {\em IEEE Conf. Comput. Vis. Pattern Recog.}, pages 4460--4470,
  2019.

\bibitem{meshry2019neural}
Moustafa Meshry, Dan~B Goldman, Sameh Khamis, Hugues Hoppe, Rohit Pandey, Noah
  Snavely, and Ricardo Martin-Brualla.
\newblock Neural rerendering in the wild.
\newblock In {\em IEEE Conf. Comput. Vis. Pattern Recog.}, pages 6878--6887,
  2019.

\bibitem{mildenhall2020nerf}
Ben Mildenhall, Pratul~P. Srinivasan, Matthew Tancik, Jonathan~T. Barron, Ravi
  Ramamoorthi, and Ren Ng.
\newblock {NeRF}: {R}epresenting scenes as neural radiance fields for view
  synthesis.
\newblock In {\em Eur. Conf. Comput. Vis.}, 2020.

\bibitem{nam2018practical}
Giljoo Nam, Joo~Ho Lee, Diego Gutierrez, and Min~H Kim.
\newblock Practical svbrdf acquisition of 3d objects with unstructured flash
  photography.
\newblock {\em ACM Trans. Graph.}, 37(6):1--12, 2018.

\bibitem{niemeyer2020differentiable}
Michael Niemeyer, Lars Mescheder, Michael Oechsle, and Andreas Geiger.
\newblock Differentiable volumetric rendering: Learning implicit 3d
  representations without 3d supervision.
\newblock In {\em IEEE Conf. Comput. Vis. Pattern Recog.}, pages 3504--3515,
  2020.

\bibitem{oechsle2020learning}
Michael Oechsle, Michael Niemeyer, Lars Mescheder, Thilo Strauss, and Andreas
  Geiger.
\newblock Learning implicit surface light fields.
\newblock {\em arXiv preprint arXiv:2003.12406}, 2020.

\bibitem{oxholm2014multiview}
Geoffrey Oxholm and Ko Nishino.
\newblock Multiview shape and reflectance from natural illumination.
\newblock In {\em IEEE Conf. Comput. Vis. Pattern Recog.}, pages 2155--2162,
  2014.

\bibitem{Park_2019_CVPR}
Jeong~Joon Park, Peter Florence, Julian Straub, Richard Newcombe, and Steven
  Lovegrove.
\newblock Deepsdf: Learning continuous signed distance functions for shape
  representation.
\newblock In {\em IEEE Conf. Comput. Vis. Pattern Recog.}, June 2019.

\bibitem{park2020seeing}
Jeong~Joon Park, Aleksander Holynski, and Steve Seitz.
\newblock Seeing the world in a bag of chips.
\newblock {\em IEEE Conf. Comput. Vis. Pattern Recog.}, 2020.

\bibitem{pytorch2019_9015}
Adam Paszke, Sam Gross, Francisco Massa, Adam Lerer, James Bradbury, Gregory
  Chanan, Trevor Killeen, Zeming Lin, Natalia Gimelshein, Luca Antiga, Alban
  Desmaison, Andreas Kopf, Edward Yang, Zachary DeVito, Martin Raison, Alykhan
  Tejani, Sasank Chilamkurthy, Benoit Steiner, Lu Fang, Junjie Bai, and Soumith
  Chintala.
\newblock {PyTorch}: An imperative style, high-performance deep learning
  library.
\newblock In H. Wallach, H. Larochelle, A. Beygelzimer, F. d\textquotesingle
  Alch\'{e}-Buc, E. Fox, and R. Garnett, editors, {\em Adv. Neural Inform.
  Process. Syst.}, pages 8024--8035. 2019.

\bibitem{ramamoorthi2001efficient}
Ravi Ramamoorthi and Pat Hanrahan.
\newblock An efficient representation for irradiance environment maps.
\newblock In {\em SIGGRAPH}, pages 497--500, 2001.

\bibitem{ravi_signal}
Ravi Ramamoorthi and Pat Hanrahan.
\newblock A signal-processing framework for reflection.
\newblock {\em ACM Trans. Graph.}, 23(4):1004–1042, Oct. 2004.

\bibitem{blind}
Fabiano Romeiro and Todd Zickler.
\newblock Blind reflectometry.
\newblock In Kostas Daniilidis, Petros Maragos, and Nikos Paragios, editors,
  {\em Computer Vision -- ECCV 2010}, pages 45--58, Berlin, Heidelberg, 2010.
  Springer Berlin Heidelberg.

\bibitem{Schmitt_2020_CVPR}
Carolin Schmitt, Simon Donne, Gernot Riegler, Vladlen Koltun, and Andreas
  Geiger.
\newblock On joint estimation of pose, geometry and svbrdf from a handheld
  scanner.
\newblock In {\em IEEE Conf. Comput. Vis. Pattern Recog.}, June 2020.

\bibitem{schoenberger2016sfm}
Johannes~Lutz Sch\"{o}nberger and Jan-Michael Frahm.
\newblock Structure-from-motion revisited.
\newblock In {\em IEEE Conf. Comput. Vis. Pattern Recog.}, 2016.

\bibitem{schoenberger2016mvs}
Johannes~Lutz Sch\"{o}nberger, Enliang Zheng, Marc Pollefeys, and Jan-Michael
  Frahm.
\newblock Pixelwise view selection for unstructured multi-view stereo.
\newblock In {\em Eur. Conf. Comput. Vis.}, 2016.

\bibitem{sitzmann2019deepvoxels}
Vincent Sitzmann, Justus Thies, Felix Heide, Matthias Nie{\ss}ner, Gordon
  Wetzstein, and Michael Zollhofer.
\newblock Deepvoxels: Learning persistent 3d feature embeddings.
\newblock In {\em IEEE Conf. Comput. Vis. Pattern Recog.}, pages 2437--2446,
  2019.

\bibitem{sitzmann2019scene}
Vincent Sitzmann, Michael Zollh{\"o}fer, and Gordon Wetzstein.
\newblock Scene representation networks: Continuous 3d-structure-aware neural
  scene representations.
\newblock In {\em Adv. Neural Inform. Process. Syst.}, pages 1119--1130, 2019.

\bibitem{tancik2020fourier}
Matthew Tancik, Pratul~P Srinivasan, Ben Mildenhall, Sara Fridovich-Keil,
  Nithin Raghavan, Utkarsh Singhal, Ravi Ramamoorthi, Jonathan~T Barron, and
  Ren Ng.
\newblock Fourier features let networks learn high frequency functions in low
  dimensional domains.
\newblock {\em arXiv preprint arXiv:2006.10739}, 2020.

\bibitem{tewari2020state}
Ayush Tewari, Ohad Fried, Justus Thies, Vincent Sitzmann, Stephen Lombardi,
  Kalyan Sunkavalli, Ricardo Martin-Brualla, Tomas Simon, Jason Saragih,
  Matthias Nie{\ss}ner, et~al.
\newblock State of the art on neural rendering.
\newblock {\em arXiv preprint arXiv:2004.03805}, 2020.

\bibitem{thies2019deferred}
Justus Thies, Michael Zollh{\"o}fer, and Matthias Nie{\ss}ner.
\newblock Deferred neural rendering: Image synthesis using neural textures.
\newblock {\em ACM Trans. Graph.}, 38(4):1--12, 2019.

\bibitem{wang2009all}
Jiaping Wang, Peiran Ren, Minmin Gong, John Snyder, and Baining Guo.
\newblock All-frequency rendering of dynamic, spatially-varying reflectance.
\newblock In {\em SIGGRAPH Asia}, pages 1--10. 2009.

\bibitem{ward1992measuring}
Gregory~J Ward.
\newblock Measuring and modeling anisotropic reflection.
\newblock In {\em SIGGRAPH}, pages 265--272, 1992.

\bibitem{wood2000surface}
Daniel~N Wood, Daniel~I Azuma, Ken Aldinger, Brian Curless, Tom Duchamp,
  David~H Salesin, and Werner Stuetzle.
\newblock Surface light fields for 3d photography.
\newblock In {\em Proceedings of the 27th annual conference on Computer
  graphics and interactive techniques}, pages 287--296, 2000.

\bibitem{xia2016recovering}
Rui Xia, Yue Dong, Pieter Peers, and Xin Tong.
\newblock Recovering shape and spatially-varying surface reflectance under
  unknown illumination.
\newblock {\em ACM Trans. Graph.}, 35(6):1--12, 2016.

\bibitem{xu2013anisotropic}
Kun Xu, Wei-Lun Sun, Zhao Dong, Dan-Yong Zhao, Run-Dong Wu, and Shi-Min Hu.
\newblock Anisotropic spherical gaussians.
\newblock {\em ACM Transactions on Graphics (TOG)}, 32(6):1--11, 2013.

\bibitem{yan2012accurate}
Ling-Qi Yan, Yahan Zhou, Kun Xu, and Rui Wang.
\newblock Accurate translucent material rendering under spherical gaussian
  lights.
\newblock {\em Computer Graphics Forum}, 31(7):2267--2276, 2012.

\bibitem{yariv2020multiview}
Lior Yariv, Yoni Kasten, Dror Moran, Meirav Galun, Matan Atzmon, Ronen Basri,
  and Yaron Lipman.
\newblock Multiview neural surface reconstruction with implicit lighting and
  material.
\newblock In {\em Adv. Neural Inform. Process. Syst.}, 2020.

\bibitem{kaizhang2020}
Kai Zhang, Gernot Riegler, Noah Snavely, and Vladlen Koltun.
\newblock Nerf++: Analyzing and improving neural radiance fields.
\newblock {\em arXiv:2010.07492}, 2020.

\bibitem{zhang2018perceptual}
Richard Zhang, Phillip Isola, Alexei~A Efros, Eli Shechtman, and Oliver Wang.
\newblock The unreasonable effectiveness of deep features as a perceptual
  metric.
\newblock In {\em IEEE Conf. Comput. Vis. Pattern Recog.}, 2018.

\bibitem{zhou2016sparse-as-possible}
Zhiming Zhou, Guojun Chen, Yue Dong, David Wipf, Yong Yu, John Snyder, and Xin
  Tong.
\newblock Sparse-as-possible svbrdf acquisition.
\newblock {\em ACM Trans. Graph.}, 35, November 2016.

\bibitem{zollhofer2015shading}
Michael Zollh{\"o}fer, Angela Dai, Matthias Innmann, Chenglei Wu, Marc
  Stamminger, Christian Theobalt, and Matthias Nie{\ss}ner.
\newblock Shading-based refinement on volumetric signed distance functions.
\newblock {\em ACM Trans. Graph.}, 34(4):1--14, 2015.

\end{thebibliography}
